\DocumentMetadata{}
\documentclass[sigplan,10pt]{acmart}
\AtBeginDocument{%
  \providecommand\BibTeX{{%
    \normalfont B\kern-0.5em{\scshape i\kern-0.25em b}\kern-0.8em\TeX}}}

\copyrightyear{2025}
\acmYear{2025}
\setcopyright{acmlicensed}\acmConference[EuroSys '25]{Twentieth European Conference on Computer Systems}{March 30-April 3, 2025}{Rotterdam, Netherlands}
\acmBooktitle{Twentieth European Conference on Computer Systems (EuroSys '25), March 30-April 3, 2025, Rotterdam, Netherlands}
\acmDOI{10.1145/3689031.3717476}
\acmISBN{979-8-4007-1196-1/2025/03}

\usepackage[linesnumbered,ruled,vlined]{algorithm2e}%[ruled,vlined]{
\usepackage[footnotesize]{subfigure}

\usepackage{fancyhdr,graphicx,amsmath,amssymb}
\usepackage{bbm}
\usepackage{enumitem}
\usepackage{titletoc}

\usepackage{tikz}

\definecolor{ao(english)}{rgb}{0.0, 0.5, 0.0}

\usepackage{changes}

\setdeletedmarkup{\color{red}{\sout{#1}}}
\setaddedmarkup{\color{blue}{#1}}
\newcommand{\quotes}[1]{``#1''}



\begin{document}

\newcommand{\aliasAPP}{VMR$^2$L\xspace}

\title{Towards VM Rescheduling Optimization Through Deep Reinforcement Learning}
\author{Xianzhong Ding}
\authornote{Both authors contributed equally to this research.}
\affiliation{%
  \institution{University of California, Merced}
}
\email{xding5@ucmerced.edu}

\author{Yunkai Zhang}
\authornotemark[1]
\affiliation{%
  \institution{University of California, Berkeley}
}
\email{yunkai_zhang@berkeley.edu}

\author{Binbin Chen}
\affiliation{%
  \institution{ByteDance}
}
\email{chenbinbin.1996@bytedance.com}

\author{Donghao Ying}
\affiliation{%
  \institution{University of California, Berkeley}
}
\email{donghaoy@berkeley.edu}

\author{Tieying Zhang}
\authornote{Corresponding author.}
\affiliation{%
  \institution{ByteDance}
}
\email{tieying.zhang@bytedance.com}

\author{Jianjun Chen}
\affiliation{%
  \institution{ByteDance}
}
\email{jianjun.chen@bytedance.com}

\author{Lei Zhang}
\affiliation{%
  \institution{ByteDance}
}
\email{zhanglei.michael@bytedance.com}

\author{Alberto Cerpa}
\affiliation{%
  \institution{University of California, Merced}
}
\email{acerpa@ucmerced.edu}

\author{Wan Du}
\authornotemark[2]
\affiliation{%
  \institution{University of California, Merced}
}
\email{wdu3@ucmerced.edu}

\begin{abstract}
Modern industry-scale data centers need to manage a large number of virtual machines (VMs). Due to the continual creation and release of VMs, many small resource fragments are scattered across physical machines (PMs). To handle these fragments, data centers periodically reschedule some VMs to alternative PMs, a practice commonly referred to as VM rescheduling. Despite the increasing importance of VM rescheduling as data centers grow in size, the problem remains understudied. We first show that, unlike most combinatorial optimization tasks, the inference time of VM rescheduling algorithms significantly influences their performance, due to dynamic VM state changes during this period. This causes existing methods to scale poorly. Therefore, we develop a reinforcement learning system for VM rescheduling, \aliasAPP, which incorporates a set of customized techniques, such as a two-stage framework that accommodates diverse constraints and workload conditions, a feature extraction module that captures relational information specific to rescheduling, as well as a risk-seeking evaluation enabling users to optimize the trade-off between latency and accuracy. We conduct extensive experiments with data from an industry-scale data center. Our results show that \aliasAPP can achieve a performance comparable to the optimal solution but with a running time of seconds. Code\footnote{The majority of work was done during the internship at ByteDance.}\footnote{\url{https://github.com/zhykoties/VMR2L_eurosys}} and datasets\footnote{\url{https://drive.google.com/drive/folders/1PfRo1cVwuhH30XhsE2Np3xqJn2GpX5qy}} are open-sourced.
\end{abstract}

\maketitle % should come after the abstract
\pagestyle{plain} % should come right after \maketitle

\begin{figure*}
    \hspace{1ex}
    \begin{minipage}[t]{0.32\linewidth}
        \centering
         \includegraphics[width=2.2in,height=0.9in,angle=0]{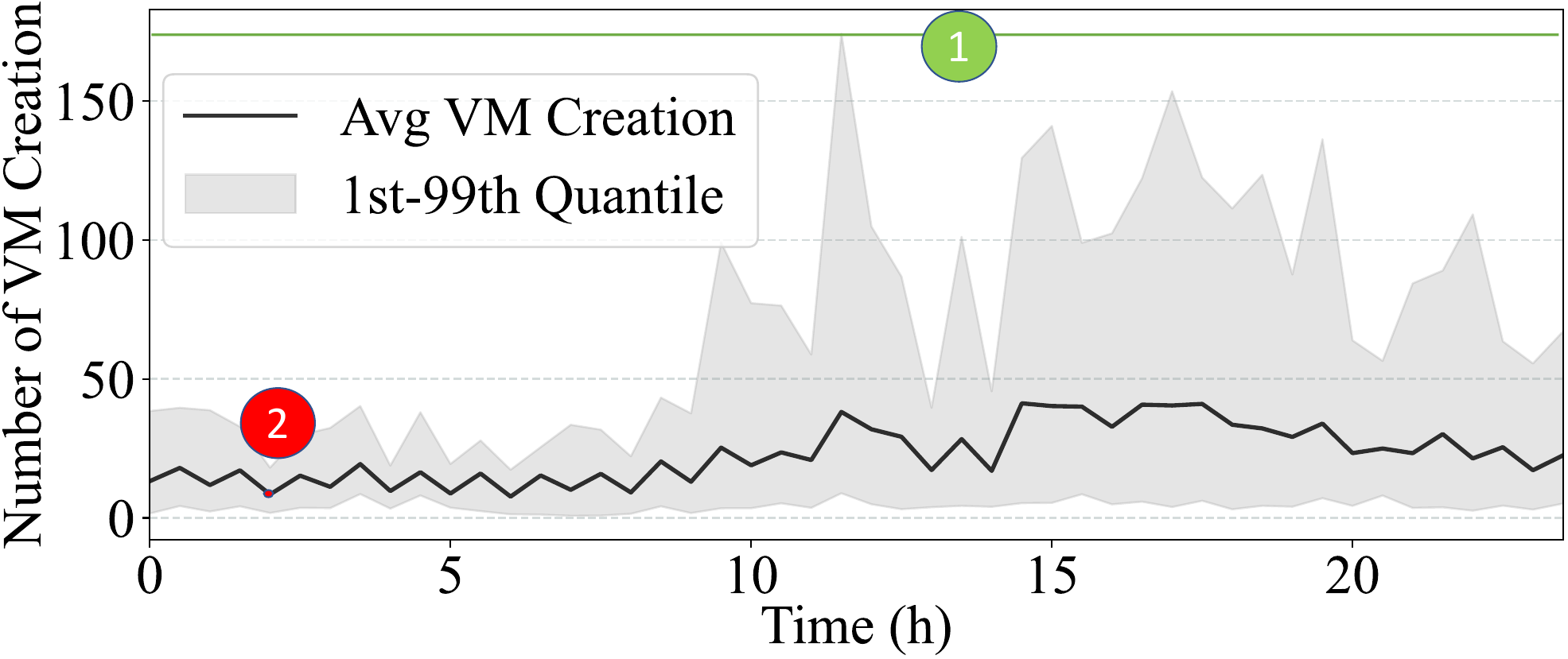}
    
        \caption{The number of VM arrivals and exits per minute. The green line indicates a continuous VMS process over 24 hours.}
        \label{fig: vm_creation}
    \end{minipage}
    \hspace{0.5ex}
    \begin{minipage}[t]{0.32\linewidth}
        \centering
         \includegraphics[width=2.2in,height=0.9in,angle=0]{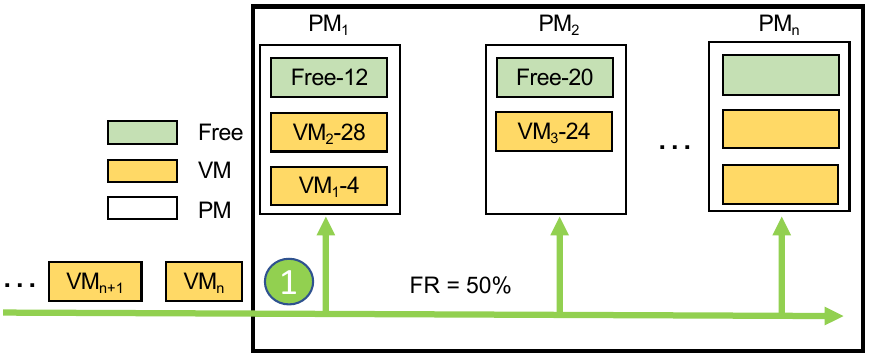}
    
        \caption{VMS process. The green number 1 denotes the VMS operation, selecting PMs for incoming VM requests.}

        \label{fig: vm_schedule_process}
    \end{minipage}
    \hspace{0.5ex}
    \begin{minipage}[t]{0.32\linewidth}
        \centering
         \includegraphics[width=2.2in,height=0.9in,angle=0]{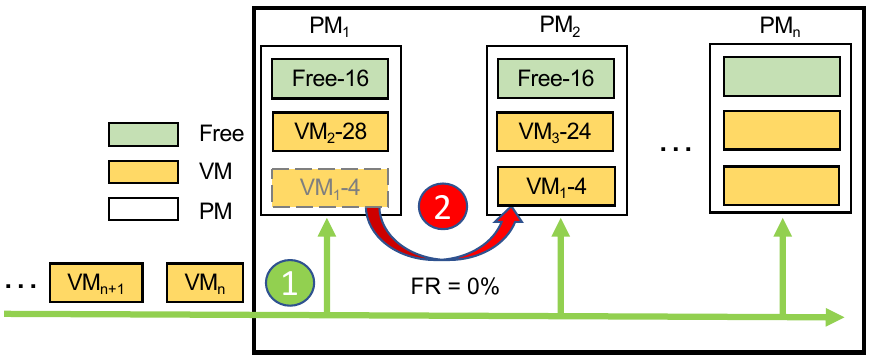}
    \caption{VMR process. The red number 2 highlights the off-peak period when VMR is typically performed.}     
        \label{fig: vm_reschedule_process}
    \end{minipage}
    % \vspace{-0.2in}
\end{figure*}

\section{Introduction}\label{sec: introduction}

Cloud service providers allow end-users to access computing resources, such as CPU and memory. They adopt resource virtualization to maximize hardware utilization, allocating Virtual Machines (VMs)~\cite{thalheim2022vmsh, clark2005live} with the requested resources to end-users. An industry-scale data center is typically organized into clusters, where each cluster has hundreds of Physical Machines (PMs), and each PM can host multiple VMs that run independently \cite{ding2023vmr2l, hadary2020protean}. 
However, if a PM already hosts several VMs and the remaining resources on the PM fail to fulfill an additional VM request, the resources leftover cannot be used are called fragments \cite{shirvani2020survey, talebian2020optimizing}. To allocate the resources efficiently, a central server manages all VM requests on PMs by performing two tasks, scheduling and rescheduling, in order to achieve different resource utilization goals, such as minimizing the overall fragment rate (FR) or minimizing the number of migrations required to achieve a specific FR.

\noindent \textbf{Fragment Rate.} FR quantifies the ratio of unusable CPU resources to total available CPU resources across all PMs. Specifically, the numerator represents the total CPU resources that cannot be used to schedule a 16-core VM (i.e., CPU fragments that are too small or scattered to accommodate such a VM). The denominator is the total available CPU resources across all PMs. This metric helps assess how efficiently the system utilizes its resources for large VM allocations.

\noindent \textbf{VM Scheduling (VMS).} When a new VM request arrives, VMS selects one PM from all available PMs that can accommodate the request. If a VM is not scheduled properly, it can directly affect the end user. 
Fig. \ref{fig: vm_creation} shows the distribution of VM changes (VMs arriving and exiting) per minute over 24 hours, averaged over a 30-day period from a cluster in our in-house data center. The y-axis represents the number of VMs changed (arrivals and exits) within each minute. To serve all users at all times, the VMS algorithm (green number 1) must handle the maximum number of VM changes, as indicated by the green line, which represents the continuous VM scheduling process throughout the day. The high queries per second (QPS) require VMS algorithms to have strict latency and stability, which deems only simple heuristic methods with short inference time feasible. In practice, ByteDance uses best-fit \cite{ha2017online, parreno2008maximal}, which sorts all PMs that meet the requirements of the current VM according to the amount of FR reduction before and after this VM is added, and chooses the PM with the largest reduction. However, such heuristic algorithms lead to many fragments scattered across PMs. Combined with the continual exiting of completed VMs, it leads to many fragments scattered across PMs. These have to be solved by VM rescheduling.

\noindent \textbf{VM Rescheduling (VMR).}
Rescheduling is critical to optimize resource usage, which migrates VMs from their current PMs to new destination PMs. Unlike VMS which needs to run throughout the day, VMR is mostly performed during off-peak hours in early mornings\footnote{In less common cases, VMR is also performed if a high FR is observed that could potentially lead to insufficient resources for upcoming VM requests.} where there are fewer VM changes as indicated by the red dot in Fig. \ref{fig: vm_creation}. Also, if a rescheduling action fails, VMs can simply stay on their original PMs without affecting the end-users. This allows the use of more advanced algorithms. 

VMR can be efficiently performed using live migration, ensuring minimal downtime. As most data centers manage VMs with compute-storage separation (i.e., using cloud disks)~\cite{KumarDSAA2022}, only the memory needs to be transferred. Specifically, we first copy the VM's memory state from the source PM to the destination PM while it continues running on the source PM. Changes to the VM’s memory during this process, known as \quotes{dirty pages} are tracked and re-copied incrementally \cite{clark2005live} until the remaining changes are small. At this point, the VM is briefly paused for a final synchronization. Since modern data centers use high-bandwidth networks for internal file transfers \cite{Wood2015Cloudnet}, this VMR process incurs a low overhead.

Note that rescheduling is primarily applied to clusters hosting VMs that use hardware virtualization, similar to Elastic Compute Cloud (EC2) environments. These VMs provide strong isolation and have a high startup cost, making them suitable for workloads running for extended periods, such as development machines \cite{VirtualHardware}. Rescheduling is unnecessary for other short-lived tasks such as CI/CD or CronJob. They are managed in separate clusters via Kubernetes, which offers fast startup through operating system-level virtualization \cite{VirtualOS}. For system stability, rescheduling is typically restricted to the same cluster. A cluster typically involves no more than several hundred PMs, since i) it allows allocation of dedicated resources to different user groups, where specific configurations can be better optimized, and ii) each cluster can be monitored and managed independently, allowing one cluster to upgrade without affecting others~\cite{cluster_design}. A migration number limit (MNL) is set to control the number of VMs to migrate and is often chosen to be $2\sim 3\%$ of all VMs.

Smaller VMs (e.g., proxy servers\footnote{A proxy server acts as an intermediary between a client requesting a resource and the server providing that resource.} or routine monitoring/testing) are easy to create using fragmented resources and have almost no risk of supply interruption. Conversely, many high-priority tasks that are directly consumer-facing require medium and large-sized VMs. Thus, our study focuses on the 16-core FR to meet the operational needs at ByteDance\footnote{A simple extension allows our framework to accommodate FR defined based other X-Core VMs under different specifications, in the form of a weighted average of several VM types, or even a combination of CPU and memory fragments. See Section \ref{sec: diff-obj}.}, where 16-core is the default VM type for development machines.

\noindent \textbf{Benefits of VMR.} Consider the FR in Fig. \ref{fig: vm_schedule_process}. PM$_1$ has 12 CPUs left and PM$_2$ has 20 CPUs left, but only PM$_2$ can host another 16-Core VM, and the remaining $12 + (20 - 16) = 16$ CPUs become fragments. The FR is therefore $16 / (12 + 20) = 50\%$. In Fig. \ref{fig: vm_reschedule_process}, VMR reassigns VM$_1$ from PM$_1$ to PM$_2$, leaving 16 free CPUs on each PM, which is just enough to handle an additional 16-Core VM. The FR after VMR becomes $0\%$.

Note that while the VMR algorithm computes a solution, VMS is still handing new VM requests and completed VMs are also being deleted. The dynamic nature of VM states causes the computed VMR solution to no longer be optimal or even feasible\footnote{A VM will not be rescheduled if it has exited or the destination PM no longer has enough resources or fails to meet other service constraints.}. Therefore, VMR also needs to be very efficient. In Section \ref{sec: motivation-exp}, we formulate VMR as a Mixed Integer Programming (MIP) problem and conduct an experiment to show that different from other MIP applications, VMR inference time must be under five seconds for the solution to remain competitive.

Most existing solutions either involve accelerating MIP solvers \cite{zhu2021network, padberg1991branch} or rely entirely on heuristics \cite{ha2017online}. However, the former still fails to meet the strict latency requirement, while the latter leads to suboptimal solutions. In this work, we develop \aliasAPP, a deep Reinforcement Learning (RL) system for VM rescheduling. 
RL is a great fit for two reasons. First, while RL often suffers from poor sample complexity \cite{Kaelbling1996}, VMR operates in a deterministic environment, meaning that, given the current state and action, the next state can be predicted exactly. This allows us to build a simulator that only requires the initial VM-PM mappings for training, without having to interact with a real data center, which \textit{drastically lowers} the number of training samples required. Second, the generalization ability \cite{mazyavkina2020reinforcement} of deep RL enables the agent to train offline and apply the learned policy directly in production \textit{without retraining}. This is crucial in meeting the strict latency requirements for VMR. We summarize the contributions of this paper as follows:

\begin{itemize}

\item \textbf{RL for VM rescheduling.} We identify the unique characteristics of the rescheduling problem in terms of latency requirement and environmental uncertainties, which motivate its formulation as an RL problem.

\item \textbf{Customized techniques for VM rescheduling.} We design i) a two-stage framework to flexibly accommodate different service constraints and address the exploration challenge, ii) a feature extraction module that scales to large data centers while capturing relational information specific to VMR, and iii) a risk-seeking evaluation pipeline that leverages the deterministic nature of VMR to offer a better trade-off between inference speed and solution quality.

\item \textbf{A \aliasAPP prototype and extensive evaluation.} We collect two real datasets and show that \aliasAPP can generalize to different objectives, service constraints, as well as abnormal workloads at deployment time. Our code and datasets are released.
\end{itemize}
\section{Motivation Experiment}\label{sec: motivation-exp}

\subsection{Problem Formulation and Two Algorithms}
\label{sec: mip_formulation}
In a data center cluster, let $\mathcal{V}, \mathcal{P}$ be the set of VMs and PMs, respectively. 
On the supply side, a PM $i \in \mathcal{P}$ is equipped with two Non-Uniform Memory Access (NUMA) nodes\footnote{A NUMA node is a subsystem within a PM that controls both memory and processing units. This architecture improves performance by allowing each node to access its local memory faster than memory on another node \cite{NUMA_architecture}.}. For PM $i$, NUMA $j$ can provide $U_{i,j}$ CPU resources and $V_{i,j}$ memory resources.
On the demand side, a VM $k \in \mathcal{V}$ requires $u_{k}$ CPU resources and $v_{k}$ memory resources and should be deployed on a single PM using $w_k \in \{1, 2\}$ NUMAs. $w_k$ is the number of NUMAs required by VM $k$ (1 for single-NUMA deployment, 2 for double-NUMA). After deploying several VMs on PM $i \in \mathcal{P}$, it remains $\tilde U_{i,j}$ spare CPU resources on NUMA $j$. We define \textbf{X-core fragment} of PM $i$ as $\sum_j (\tilde U_{i,j} \% X)$, i.e., the remaining CPUs cannot be further utilized by additional X-core VMs.

Given $M$ VMs that are initially assigned to $N$ PMs, VMR reassigns a subset of deployed VMs and migrates them onto some new PMs. The \textit{Migration Number Limit (MNL)} is a tunable parameter that defines the maximum number of VMs that can be migrated during each rescheduling task. By adjusting MNL, we can control the trade-off between performance improvements and migration overhead. We formulate VM rescheduling as an optimization problem that searches for the optimal reassignment of VMs (up to the specified MNL), with the goal of minimizing the total X-core fragments across all PMs:

\begin{table*}
  \caption{VM types considered in the main experiments. Extra resource and service constraints are considered in Section \ref{sec:evaluation}.}
  \label{tab: vmtype}
  \small
  \centering
  \renewcommand{\arraystretch}{1} % Reduce line spacing to 80% of normal
  \begin{tabular}{llllllll}
    \hline
     VM Types& large  & xlarge   & 2xlarge &4xlarge & 8xlarge & 16xlarge &22xlarge\\
     \hline
     Requested CPU & 2 & 4 & 8 & 16 & 32 & 64 & 88\\
     Requested Memory (GB) &4&8&16&32&64&128&176\\
     Deploy NUMA & Single & Single & Single & Single & Double & Double & Double\\
    \hline
  \end{tabular}
\end{table*}

\begin{align}
\vspace*{-\baselineskip}
   &\text{Minimize: }& \sum_{i,j} \left ( 
        U_{i, j} - \sum_k \frac{x_{k,i,j}\cdot u_k}{w_k} - Xy_{i,j}
   \right ) 
  \label{eq: fr_goal}\\
   & \text{Subject to:} &\sum_k \frac{x_{k, i, j}\cdot u_{k}}{w_{k}} + Xy_{i,j} \le U_{i,j}, \label{eq: res-cpu}\\
    &&\sum_k \frac{x_{k, i, j} \cdot v_{k}}{w_{k}} \le V_{i,j}, \label{eq: res-mem}\\
    &&\sum_{i,j} x_{k,i,j} = w_k, \label{eq: deploy}\\
    &&\sum_k (1 - x_{k, i_k, j_k}) \le MNL, \label{eq: mig-num}\\
    &&x_{k,i,0} = x_{k,i,1}, \quad \forall k \in \{k|w_k=2\}, \label{eq: double-numa}\\
    &&x_{k,i,j} \in \{0,1\}\text{ and } y_{i,j} \in \mathbb{Z}.
  \label{constraint}
\end{align}
Here, $\{x, y\}$ are the decision variables, where $x_{k,i,j}$ represents whether VM $k$ is deployed to the NUMA $j$ of PM $i$ in the new assignment (0 for No, 1 for Yes), and $y_{i, j}$ represents the maximum number of X-core VMs can be deployed on NUMA $j$ of PM $i$ using the remaining CPU resources. The objective in Equation \ref{eq: fr_goal} is to minimize the total X-core fragments.

Equation \ref{eq: res-cpu} and \ref{eq: res-mem} enforce that the resource usage by VMs cannot exceed the total capacity of a PM. Equation~\ref{eq: deploy} indicates that each VM must be deployed on exactly one PM. Equation~\ref{eq: mig-num}, in which $i_k$ and $j_k$ are the initial PM id and NUMA id (0 for double-NUMA VMs) of VM $k$, means the total migration number should not exceed the limit. Lastly, Equation~\ref{eq: double-numa} forces VMs with double NUMAs to deploy both NUMAs on the same PM. 

Note (1) each PM has two NUMAs; (2) $w_k$ is a constant for each VM as determined by their types (Table \ref{tab: vmtype}). Thus, $\sum_{i,j} x_{k,i,j} = w_k$ (Equation~\ref{eq: deploy}) enforces that the actual NUMA allocation number of VM $k$ matches the desired configuration. When $w_k = 1$, Equation~\ref{eq: deploy} constraints VM $k$ to be deployed on one NUMA of a PM; when $w_k = 2$, Equation~\ref{eq: deploy} constraints VM $k$ to be deployed on both NUMAs of a PM. Note that deploying VM $k$ on two NUMAs of two different PMs (each PM hosting a NUMA) violates $x_{k,i,0} = x_{k,i,1}, \forall k \in \{k|w_k=2\}$ (Equation~\ref{eq: double-numa}). Because $w_k \neq 0$, it guarantees each VM is deployed.

\noindent \textbf{Mixed Integer Programming (MIP) Solvers.} The above optimization problem can be solved by an off-the-shelf MIP solver such as CPLEX \cite{CPLEX} and Gurobi \cite{Gurobi_solver}, which finds a near-optimal solution through branch \& bound, cutting planes, etc. In our experiments, we use Gurobi. 

\noindent \textbf{Heuristic Algorithm (HA).} \label{greedy_algorithm} To obtain a feasible solution within a short time frame, heuristic algorithms are often used in industry data centers \cite{kubernetes_scheduler}. They normally include two stages: filtering and scoring. In the filtering stage, we calculate the change in FR for each VM as if it is removed from its source PM, and only select the VM candidate that corresponds to the most drop in FR. In the scoring stage, we calculate the change in FR as if the selected VM is migrated to each of the eligible PMs. We then greedily assign the selected VM to the PM that leads to the largest drop in FR. The above two stages are repeated until the MNL is reached.

\begin{figure}[t]
\centering
\subfigure{
\includegraphics[width=1.5in,height=1.23in]{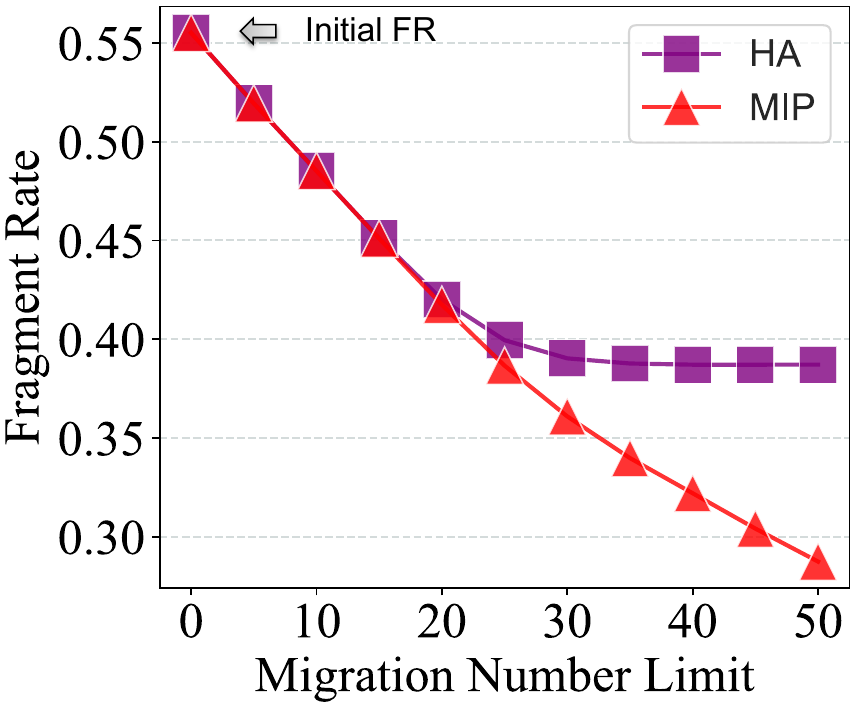}
\label{fig: fr_mip_heurstic}

}
\subfigure{
\includegraphics[width=1.5in,height=1.2in]{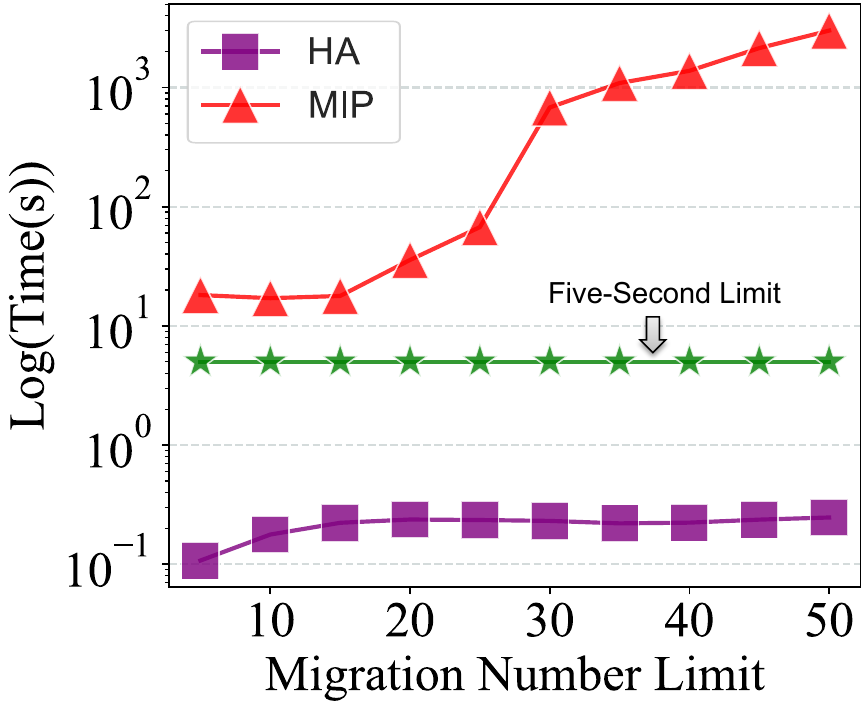}
\label{fig: time_mip_heurstic}
}
   \vspace{-0.2in}
\caption{FR and inference time at different MNLs.}
\label{fig: motivation_exp}
 \vspace{-0.5cm}
\end{figure}

\begin{figure}[t]
\begin{center}
  \includegraphics[height=1.1in, width=3.0in]{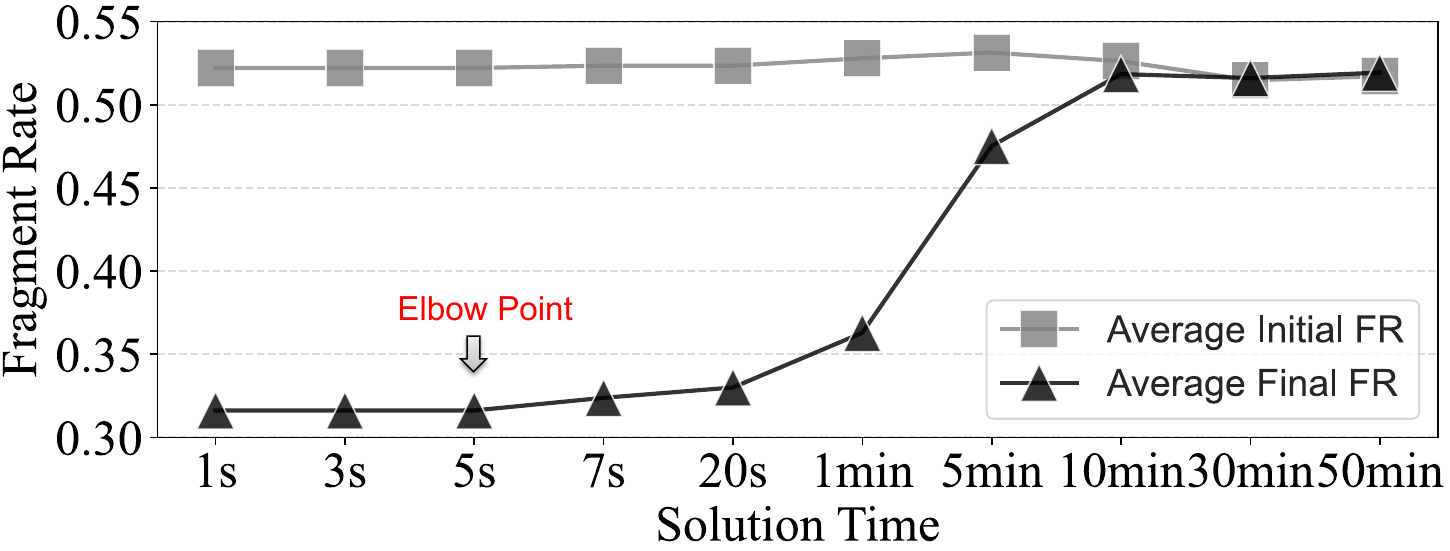}
  \end{center}
  \vspace{-0.2in}
  \caption{Effect of inference time on achieved performance.}
  \label{fig: performance_effect}
\end{figure}

\subsection{Experiment Results}
We conduct experiments to quantify the performance of the above MIP and HA in terms of FR and inference time.
We use a real dataset from a data center with 280 PMs and 2089 VMs. The detailed experiment settings can be found in Section \ref{sec: implementation}. 

Fig. \ref{fig: motivation_exp} (left) depicts the FR performance of MIP and HA under different MNLs. 
MIP's FR is lower than HA's since it guarantees a near-optimal solution. 
Moreover, the FR gap between MIP and HA becomes larger as MNL increases, because HA only exploits the action that would lead to the most significant drop in FR when it migrates one VM. After migrating 25 VMs, the heuristic algorithm can no longer find any more VMs that can lower the FR. In Fig. \ref{fig: motivation_exp} (right), we see that as the MNL increases from 25 to 50, the computation time of MIP grows exponentially, taking 1.78 minutes for 25 migrations and 50.55 minutes for 50 migrations. This poor time complexity is unacceptable in data centers where new VM requests continually come in, and the problem state is constantly changing. 

To see how a near-optimal solution can result in a suboptimal achieved performance due to its poor inference time, we conduct an experiment on real traces from our in-house data center by selecting 200 random initial VM-PM mappings. For each mapping, we use Gurobi to compute a near-optimal solution to the MIP formulation of VMR, which takes 50.55 minutes. However, since VMs were dynamically arriving and exiting, most actions were no longer feasible and would fail to be deployed after 50 minutes. We then compute the final performance that could be achieved as if the near-optimal solution was instead returned in a shorter period of time, averaged over the 20 mappings. Fig. \ref{fig: performance_effect} shows that the solution remains near-optimal if it could be computed within five seconds, as highlighted by the \quotes{elbow point}. However, FR reduction quickly diminishes once the inference time exceeds that. Therefore, we require all methods to be able to return a solution within a \textbf{\textit{five-second limit}} for each mapping during inference (green line in Fig. \ref{fig: motivation_exp} (right)).

\noindent \textbf{Limitations of Current Methods.}
From the above experiment, we see that the primary issue of the MIP approach is its poor time complexity, which prevents it from scaling to industry-scale data centers with thousands of PMs and VMs. To reduce the execution time of MIP solvers, some current methods rely on estimating feasible solutions using proprietary heuristic methods and then using branch-and-cut techniques \cite{padberg1991branch} to identify optimal solutions. The hand-tuned heuristics are based on human expertise to overcome the scalability challenge by pruning the search space. Yet, the heuristics have to be designed separately for different cluster conditions in every data center as there are no universal heuristics for all scenarios. For this reason, at ByteDance, we use Partitioned Optimization Problems (POP) \cite{narayanan2021solving} that randomly partition the rescheduling problem into several subproblems and apply MIP to each subproblem. However, as we show in Section \ref{sec:baselines}, POP performs suboptimally under the five-second limit. Instead, we aim to design a solution that can match the performance of MIP, meet the latency requirements, and does not require manual feature engineering for different clusters.

\section{Design of \aliasAPP}\label{sec: new_design}
\subsection{VM Rescheduling as an RL Problem}\label{sec: vmr-as-rl}

Deep RL has demonstrated remarkable abilities in many domains \cite{silver2017mastering, ding2024exploring}. Without requiring pre-programmed heuristic rules from experts, deep RL \textit{learns directly from data}, but its main drawback is the amount of training data required \cite{ding2025safe, an2024go}. Additionally, rescheduling MNL VMs simultaneously requires an action space of $O((M \cdot N)^{MNL})$, which is too large for the agent to learn effectively.

To address these challenges, we formulate the problem such that a VMR request starts an episode, which involves MNL steps. At each migration step, the action of the agent (\aliasAPP) reschedules a single VM from its source PM to a new destination PM based on the current PM and VM status. Notably, the environment is deterministic -- given the current state and action, we can exactly know the next state and the change in objective. 

To this end, we design a simulator following the OpenAI Gym environments \cite{openai} that allows us to train \aliasAPP offline --- we collect training samples from the data center, where each sample is the status of all VMs and PMs when a VMR request is created. Each sample serves as the initial state of an episode. For each step in the episode, given the current state and the agent's action, the simulator computes i) the next state and ii) a reward based on the change in the rescheduling objective, without the need to interact with the actual data center. The agent in turn uses this reward signal to gradually improve the rescheduling policy. At deployment time, using neural networks as feature extractors allows \aliasAPP to generalize to VM-PM mappings not encountered during training without retraining or finetuning.

Given the above design, we now formalize the state, action, and reward of the \aliasAPP framework.

\noindent
\textbf{State Representation.} At each step, the state serves as input to the agent, which is parameterized using neural networks. The state input contains two sets of features. The first set is the PM features, which contain four features for each of the two NUMAs of each PM, specifically the remaining CPU and memory resources, current FR, and fragment sizes. The second set is the 14 VM features, which include requested CPU and memory for each NUMA, fragment sizes, concatenated with the source PM features. If a single NUMA is requested, zeros are used as placeholders for the other NUMA. Each feature dimension is min-max normalized.

\noindent
\textbf{Action Representation.} Given the state at each step, the agent outputs an action to migrate one VM. The action at each step can be represented as a 2-tuple $(k, i)$, which means to reschedule a VM $k \in \mathcal{V}$ from its source PM to a destination PM $i \in \mathcal{P}$. Note that the source PM can be retrieved once we select $k$, so we do not include it in the action space.

\noindent
\textbf{Reward Representation.} Reward represents the immediate evaluation of the benefits each migration step brings under the given state. The goal of VM rescheduling is to minimize the FR across all PMs. While we could return the FR of all PMs as a single final reward to the agent after finishing an entire episode, it corresponds to a form of sparse reward which is known to be difficult for training \cite{rengarajan2022reinforcement}, as the agent cannot easily attribute the drop in FR to a certain migration step. Instead, we propose to generate dense rewards and use the change in fragment sizes on the source PM and the destination PM as an intermediate reward at each step. Since we focus on 16-core CPUs, the maximum change in fragment size on a single NUMA due to adding or removing a VM is $-15$ to $15$. We normalize the reward by dividing it with a constant $c=64$ so that its range is naturally scaled down to $[-\frac{15 \times 4}{64}, \frac{15 \times 4}{64}]$~\cite{henderson2018deep}. We calculate the rescaled fragment size on both NUMAs by 
\begin{equation}
    S_i = \sum_{j=0}^{1} \left( \tilde{U}_{i,j} \% X \right) \div c,
\end{equation}
and define reward as 
\begin{equation} \label{eq: original-reward}
    R = (S_{\text{before, src}} - S_{\text{after, src}}) + (S_{\text{before, dest}} - S_{\text{after, dest}}),
\end{equation}where $S_{\text{before}, \cdot}$ and $S_{\text{after}, \cdot}$ are fragment changes before and after the selected VM leaves (enters) the source (destination) PM.

\begin{figure}[t]
\begin{center}
  \includegraphics[width=1.0\columnwidth]{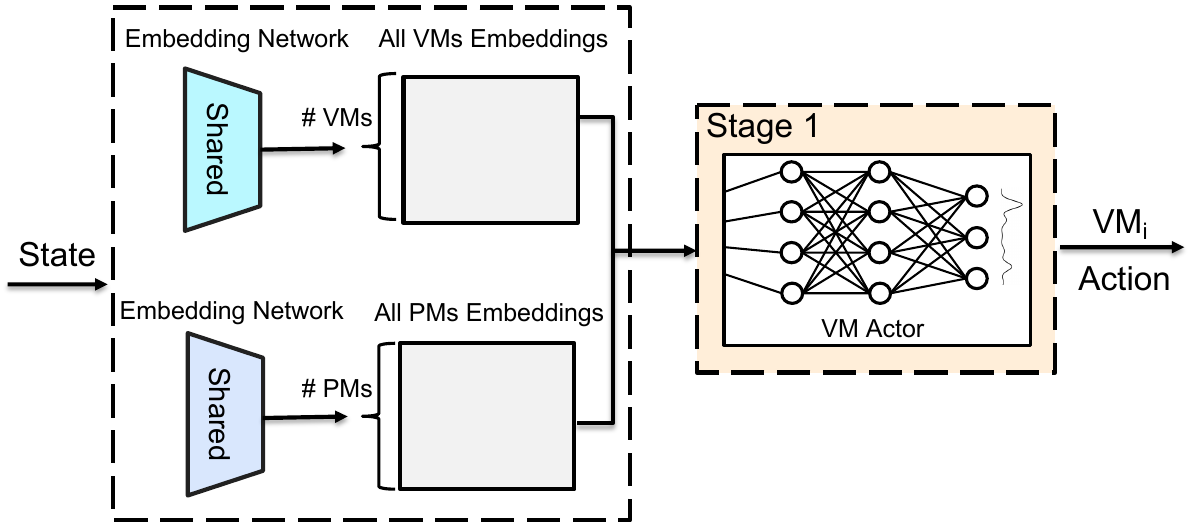}
  \end{center}
    \vspace{-0.2in}
  \caption{The first stage of \aliasAPP processes all VMs and PMs via shared embedding networks, based on which the VM actor selects a VM to be rescheduled.}
  \label{first_stage}
  % \vspace{-0.5cm}
\end{figure}

\begin{figure}[t]
\begin{center}
  \includegraphics[width=1.0\columnwidth]{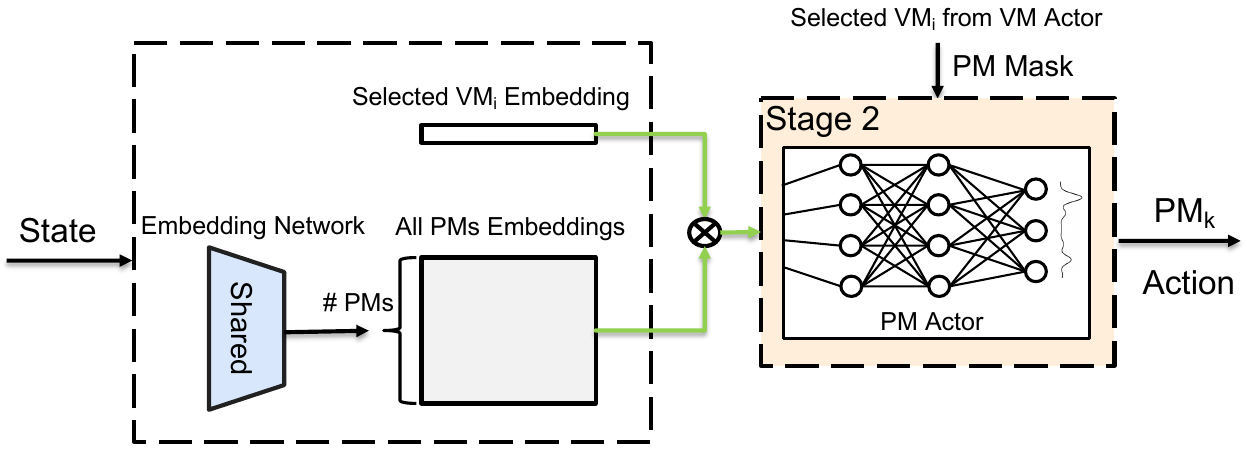}
  \end{center}
    \vspace{-0.2in}
  \caption{Once a candidate VM is selected by the VM actor, \aliasAPP masks out all the PMs that cannot host the candidate VM. The PM actor only accesses the selected VM, and then selects a destination PM from the unmasked PMs.}
  \label{second_stage}
  \vspace{-0.5cm}
\end{figure}

\subsection{Two-Stage Framework}\label{sec: two-stage-framework}
When the RL agent chooses to reschedule a VM $k$ from its source PM to a destination PM $i$, PM $i$ must have enough available CPU and memory to host VM $k$. In practical scenarios, we may also consider additional constraints to ensure service stability. For example, an application may require some VMs to be hosted across different PMs, which can be enforced in the form of a hard anti-affinity constraint.

Off-the-shelf RL models impose such hard constraints typically by invoking a heavy penalty if an illegal action is chosen, or by directly setting the probabilities for all illegal actions to be zero. As shown in Section \ref{sec: diff-constraint}, heavy penalties can result in gradient instability and lead to an inferior convergence rate. Furthermore, as the size of action space is $O(M \cdot N)$, masking all illegal actions is overly time-consuming. 

To better accommodate a variety of constraints, we leverage the characteristics of VMR and design a two-stage framework that allows the action tuple to be generated sequentially. In Stage 1 (Fig. \ref{first_stage}), the VM actor selects the VM candidate to be rescheduled. Once a candidate VM is selected, we can efficiently mask out all the PMs that cannot host the candidate VM. We then proceed to Stage 2 (Fig. \ref{second_stage}), where the PM actor selects an appropriate destination PM from the remaining PMs. Such a framework has three benefits. First, it completely avoids illegal actions for various types of constraints and thus circumvents the necessity of heavy penalties. Second, it dedicates two separate networks to select the VM candidate and the destination PM, which simplifies the VMR task by decomposing the action tuple. Finally, when we select a VM to reschedule, a considerable portion of the PMs cannot meet its resource requirements. The proposed framework can avoid exploration on these PMs and thus mitigate the exploration challenge.

\subsection{Feature Extraction with Sparse Attention}\label{sec: sparse-attention}
\textbf{Scale to Many VMs \& PMs.} For effective rescheduling decisions, \aliasAPP must extract meaningful representations of the state observation, which include features of each individual PM and VM as well as their affiliations. As Fig. \ref{fig: vm_creation} implies, the number of VMs can vary drastically even in the same cluster. This implies that the size of the features at each time step is also highly dynamic. To encode these features, one option is to concatenate features from all VMs and PMs into a long vector. However, this approach cannot handle an arbitrary number of VMs as neural networks usually require fixed-sized inputs, and it also requires a model with many parameters that would be difficult to train. 

Instead, we propose to share two small embedding networks across all VMs and PMs --- one to process each PM's features and another one to process each VM's features (Fig. \ref{first_stage} and \ref{second_stage}). As such, the number of weight parameters is \textit{independent} of the number of machines in the system. This is achieved via an attention-based transformer model \cite{vaswani2017attention, chen2024marlp, zhang2025survey, hong2024next} but tailored for rescheduling. Transformers have demonstrated strong performances in Natural Language Processing \cite{Bert}, Computer Vision \cite{VisTrans}, as well as combinatorial optimization, such as in vector bin-packing \cite{Zhang2021Attend2PackBP, li2020solving}. However, compared to bin-packing, there is a notable difference in VM rescheduling: we must choose from a set of VMs that have already been assigned to PMs.

\begin{figure}[t]
\begin{center}
  \includegraphics[height=1.3in, width=3.2in]{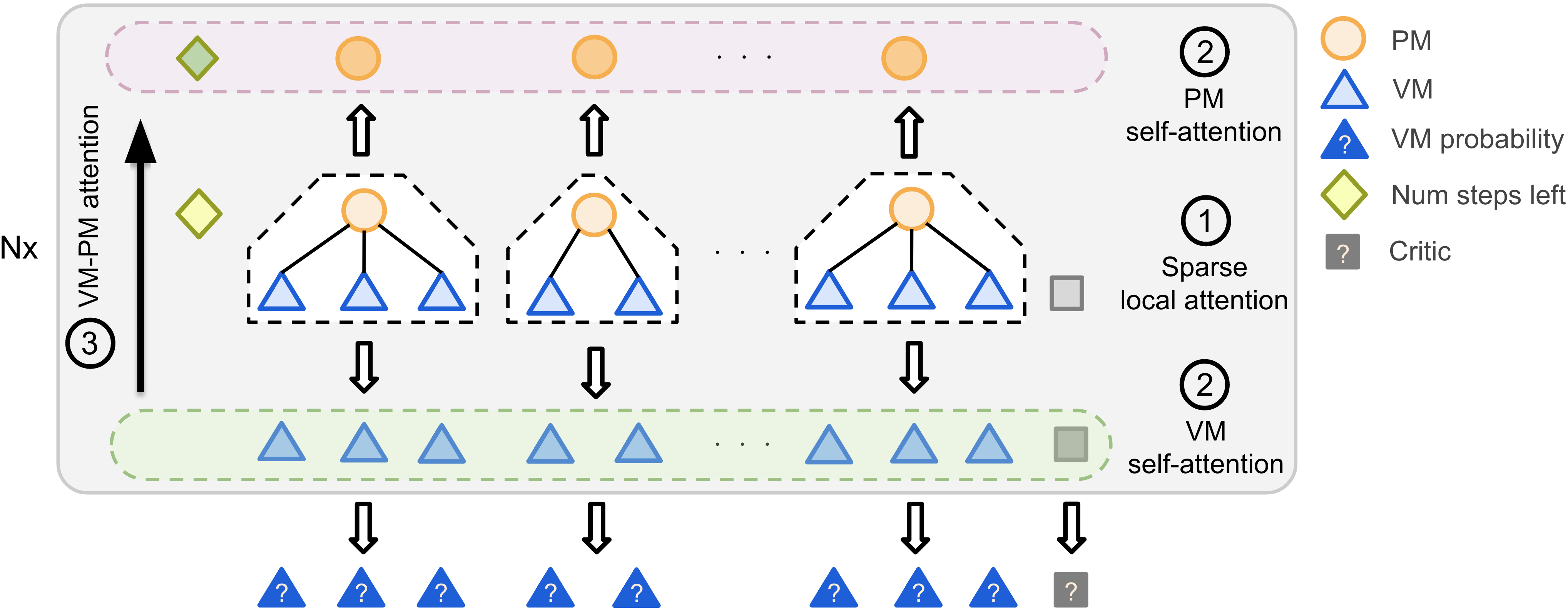}
  \end{center}
  \vspace{-0.2in}
  \caption{VM actor architecture with sparse local-attention to capture the tree-level features.}
  \vspace{-0.2in}
  \label{fig: neural_arch}
\end{figure}

\noindent \textbf{Tree-level Features.} Consider a PM with 2 CPUs left. It contains a VM with 4 CPUs and another VM with 2 CPUs. Suppose a second PM has a fragment size of 8 while hosting a VM with 8 CPUs. To minimize the total 16-core fragments, an ideal approach would be to first remove the two VMs with 2 and 4 CPUs from the first PM, and then reassign the VM with 8 CPUs from the second PM to the first PM. However, if we merely include the source PM's features in each of the VM's features and feed them into the vanilla transformer model, there will not be sufficient information for the two actors to take the above actions. Instead, each VM must be aware of the other VMs that are hosted on the same PM, which is not possible in the vanilla transformer model. In fact, each PM can be viewed as a tree of depth one, where the PM acts as the root node and the VMs it hosts act as the leaf nodes. In order to allow every VM to recognize which other VMs are hosted on the same PM, we propose to include an additional stage of \textit{sparse local-attention} within each PM tree, i.e., we only allow PMs and VMs to attend to each other if and only if they belong to the same tree. \\
\\\textbf{Architecture Overview.} We modify the vanilla transformer architecture as follows. The model is composed of several attention blocks, where each block includes three stages as shown in Fig. \ref{fig: neural_arch}:
\setlist{nolistsep}
\begin{enumerate}[noitemsep]
    \item All PMs and VMs exchange information if they belong to the same tree via sparse local-attention.
    \item Each PM attends to other PMs' updated embeddings and each VM attends to other VMs' updated embeddings with self-attention.
    \item The new VM embeddings attend to the new PM embeddings through VM-PM attention.
\end{enumerate}
After the three stages, each machine is further processed by two dense layers and layer norm \cite{layernorm}. The updated embeddings are then fed into the next block and the process repeats. Finally, the VM embeddings from the last block are linearly projected into a set of logits followed by Softmax \cite{bishop2007} to generate the probability of selecting each VM.

As for the PM actor, we adapt the vanilla encoder-decoder transformer, since we can directly inject the relational information by including the updated VM and PM embeddings from the VM actor as input. We only feed in the selected VM candidate to the encoder, while the decoder still takes in all PMs. Additionally, we also add the VM-PM attention score from stage 3 for the selected VM, since the score indicates which PMs the VM actor attends to and encourages the two actors to better coordinate. The output embeddings of each PM is linearly projected into a logit. Based on the selected VM, we mask out all the illegal PMs by setting their logits to be $-\infty$. The remaining logits are translated into the probability of selecting each PM as the destination PM.

\subsection{Risk-seeking Evaluation}\label{sec: risk-seeking-eval}
VMR is distinct from general RL problems as it allows access to a perfect world model with no uncertainties from the environment. In other words, the simulator can directly compute the final state and the corresponding objective value for a given initial state and action sequence. 

To take advantage of this distinction, we propose \textit{risk-seeking evaluation}, which is to \textbf{sample multiple trajectories during policy evaluation, and only deploy the one with the highest reward}. Given a trained \aliasAPP checkpoint, to obtain varied trajectories during inference, we sample actions from the learned policy, $\pi(\cdot \vert s)$, rather than exclusively selecting the action with the highest probability. Note that different migration trajectories can be generated in parallel if multiple GPUs are available, without significantly affecting the inference time. It is also worth mentioning that the concept of using the best-performing trajectory can be further extended to the training process, known as \textit{risk-seeking training} \cite{petersen2021deep}.

\textbf{Action Thresholding.}  It is important to note that the learned policy is likely to differ from the optimal policy due to approximation errors. In conventional RL applications, this might not be an issue since only the action assigned with the highest probability is chosen, allowing us to safely ignore the approximation errors. However, in VMR, we must sample actions from $\pi(\cdot \vert s)$ in order to sample multiple trajectories. Suppose actions with probability less than a threshold $\epsilon$ are sub-optimal. Although these actions may not be executed in a single period, they are likely to be performed at some point in the entire trajectory. 

Let $p_1 = \min_{s\in S} \sum_{a\in A} \pi_\theta(a\vert s)\cdot \mathbbm{1}\{\pi_\theta(a\vert s)\leq \epsilon \}$ be the lowest total probability of sub-optimal actions over all states. Then, the probability that the agent does not perform any sub-optimal actions along $MNL=50$ rescheduling steps is upper bounded by $(1-p_1)^{MNL}\leq e^{-MNL \cdot p_1}$. If $p_1 = 0.005$, then $23\%$ of the trajectories we sample will contain sub-optimal actions. Therefore, at each migration step, the VM actor computes the probability of selecting each VM candidate, and we calculate a threshold based on the quantile of all VM probabilities. We then mask out all VM candidates that are assigned with probabilities falling below the threshold, and similarly for PMs.

\section{Implementation}\label{sec: implementation}

\noindent \textbf{Datasets.} We collect two datasets\footnote{While a data center might have over 10,000 PMs, they are typically organized into clusters, each consisting of hundreds of PMs, for improved fault isolation and easier management. Migrations usually occur within each cluster. Our Large dataset is already larger than the size of a typical cluster. See Section \ref{sec: introduction}.} from real industry-scaled data centers -- a Medium dataset with up to 2089 VMs and 280 PMs, and a Large dataset with up to 4546 VMs and 1176 PMs\footnote{Note that the Large dataset has a lower VM to PM ratio, since it has larger average VM sizes. The datasets are collected from real clusters, which are managed by different service teams. This reflects the operational needs of different services.}. Each dataset contains 4400 mappings (or instances), which represent the assignments of VMs to PMs at various points in time. To release these datasets while ensuring confidentiality of business operations and avoiding potential train/test leakage, we anonymize each mapping by randomly removing some of the existing VMs and redeploy the VMs to any PMs that they can fit.
We split the 4400 mappings into 4000 for training, 200 for validation, and 200 for testing.
Our designed Gym simulator directly supports the dataset format.
To our knowledge, these are the largest datasets for VM rescheduling based on real traces, and shall be very useful to the community.

\noindent \textbf{Algorithm Specifics.} We implement \aliasAPP based on the CleanRL framework~\cite{huang2022cleanrl} using PPO as the backbone \cite{schulman2017proximal}. \aliasAPP contains about 8.5K lines of Python code. The framework is implemented using PyTorch \cite{pytorch, yang2023link}. The number of model parameters is independent of the number of VMs or PMs, allowing it to scale to large data centers. \aliasAPP is lightweight -- the saved checkpoint has a size less than 2 MB.

\noindent \textbf{Experiment Setup.} All models are trained on a Linux server using eight CPU cores (Intel Xeon E5) and one GPU (NVIDIA RTX 3090) \cite{lin2023drgpum, lin2025understanding}. \aliasAPP 
takes 92 hours to train, and 1.1 seconds to solve each mapping.
We report the average over 3-5 runs with different random seeds and show the confidence intervals in the convergence plots.
\section{Evaluation}
\label{sec:evaluation}

We conduct extensive experiments to answer:

\noindent
\textbullet~~How far is \aliasAPP from the optimal solution? ($\S $ \ref{sec: fr_time_performance})

\noindent
\textbullet~~How much gain does each component provide? ($\S $ \ref{sec: component})

\noindent
\textbullet~~How does the two-stage framework allow \aliasAPP to accommodate different constraints: i) constraints on the original Medium dataset, ii) multi-dimensional resource constraints, and iii) service affinity constraints? ($\S $ \ref{sec: diff-constraint})

\noindent
\textbullet~~Can \aliasAPP optimize i) an objective other than FR, ii) a mixed objective defined with multiple resource types? ($\S $ \ref{sec: diff-obj})

\noindent
\textbullet~~How well does \aliasAPP generalize to i) workloads different from the train distribution, ii) MNLs that are different at inference time, iii) more PMs and VMs, iv) different workloads with different MNLs, and (v) varying workloads with different MNLs?($\S $ \ref{sec: generalization})

\noindent
\textbullet~~Is a larger cluster more difficult for \aliasAPP to learn? ($\S $ \ref{sec: larger_cluster})

\noindent
\textbullet~~Where do the improvements come from intuitively? ($\S $ \ref{sec: case-study})

\begin{figure*}[t]
\centering
\subfigure{
\includegraphics[width=3.2in,height=1.5in]{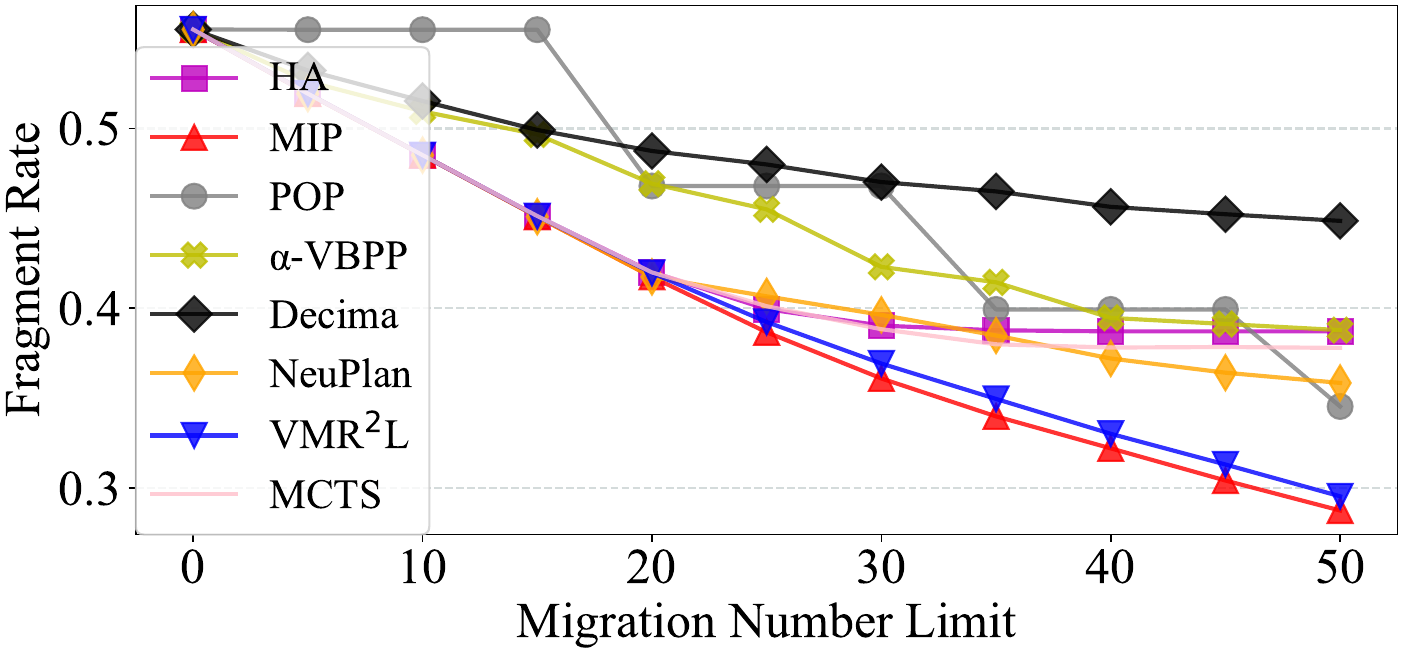}
\label{fig: fr_three_methods}
}
%\quad
\subfigure{
\includegraphics[width=3.2in,height=1.5in]{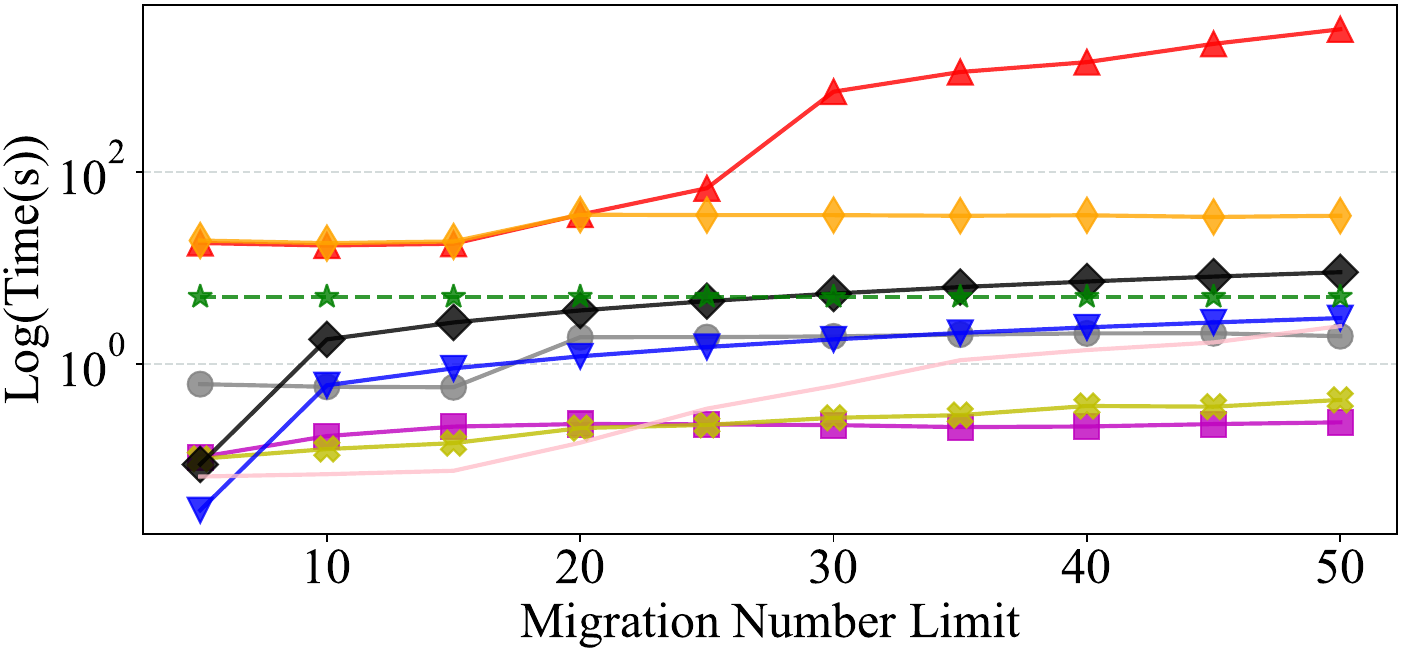}
\label{fig: time_three_methods}
}
\vspace{-0.2in}
\caption{Fragment rate (left) and inference time (right) of \aliasAPP compared with baselines at different MNLs.}
\label{fig: motivation_exp_rl}
\end{figure*}

\subsection{Existing Baseline Algorithms}\label{sec:baselines}
As later discussed in Section \ref{sec: related-work}, existing baselines can be summarized into six categories: heuristics (e.g., filtering-based heuristic, $\alpha$-VBPP), optimization algorithms (e.g., MIP), approximate algorithms (e.g., POP), search-based algorithms (e.g., MCTS), deep learning-based (e.g., Decima), and hybrid methods (e.g., NeuPlan). We compare with at least one representative algorithm from each category.

\noindent \textbf{MIP Algorithm:} introduced in Section \ref{sec: mip_formulation}. 

\noindent \textbf{Filtering-Based Heuristic Algorithm (HA):} introduced in Section \ref{greedy_algorithm}.

\noindent \textbf{Vector Bin Packing Problem ($\alpha$-VBPP):} We generalize the VBPP \cite{panigrahy2011heuristics} algorithm for initial scheduling to rescheduling. We first divide the entire episode into $MNL/\alpha$ stages. During each stage, we greedily remove $\alpha$ number of VMs that lead to the most fragments, and then apply VBPP to treat them as incoming VMs. We carefully tune $\alpha$ (10 in our case) to achieve the best FR reduction.

\noindent \textbf{Partitioned Optimization Problems (POP) \cite{narayanan2021solving}:} It solves the optimization problem formulated in Section \ref{sec: mip_formulation} by randomly splitting the problem into subproblems (each containing a subset of VMs and PMs), applying an MIP solver to each subproblem, and finally combining the results into a global solution. 
We choose the smallest number of subproblems (16 in our case) that allows POP to meet the five-second limit.

 \begin{figure*}[h]
		\hspace{1ex}
		\begin{minipage}[t]{0.32\linewidth}
		\centering
		 \includegraphics[width=2.2in,height=1.2in,angle=0]{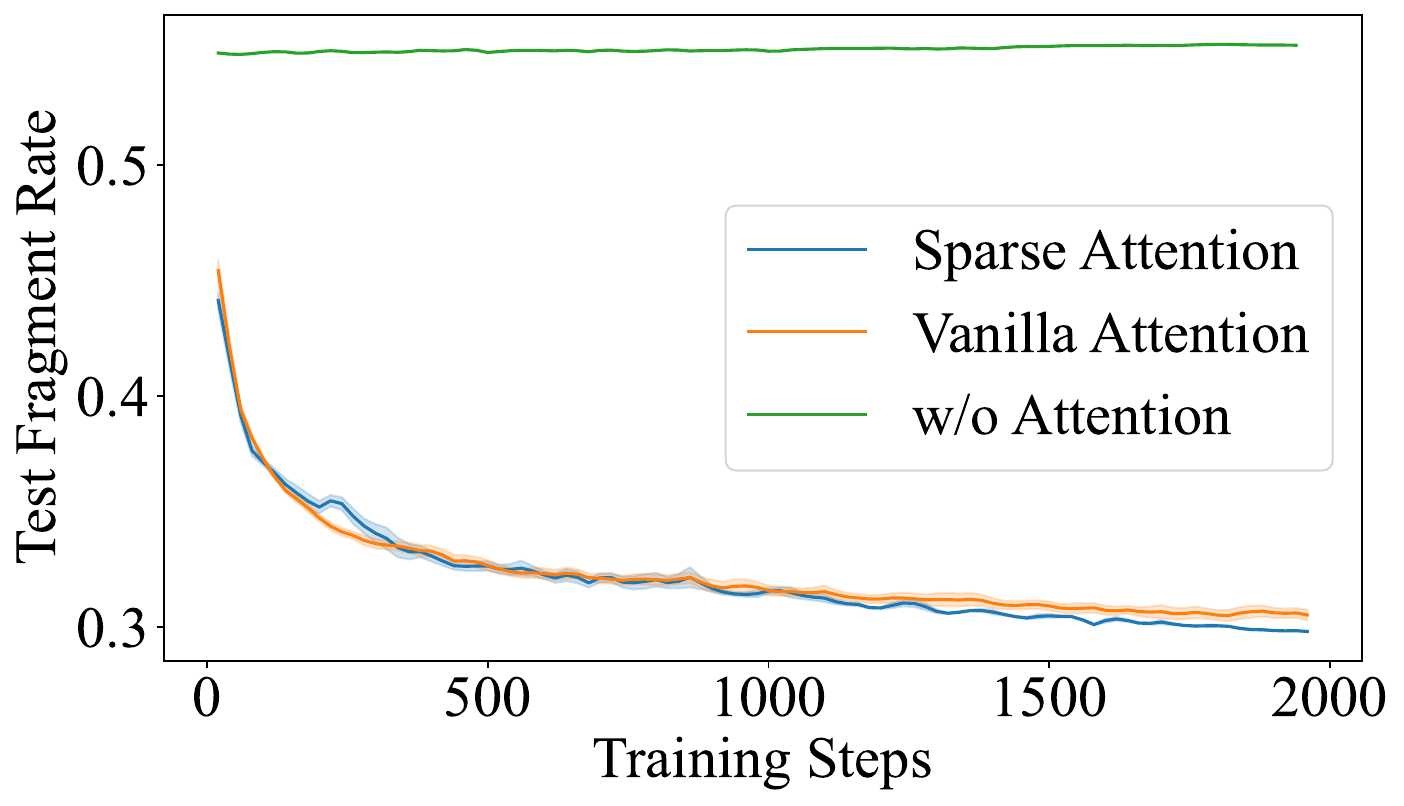}
            \vspace{-0.1in}
		\caption{Ablation on Sparse Attention.}
		\label{fig: ablation-sparse-attn}
	\end{minipage}
	\hspace{0.5ex}
		\begin{minipage}[t]{0.32\linewidth}
		\centering
		 \includegraphics[width=2.2in,height=1.2in,angle=0]{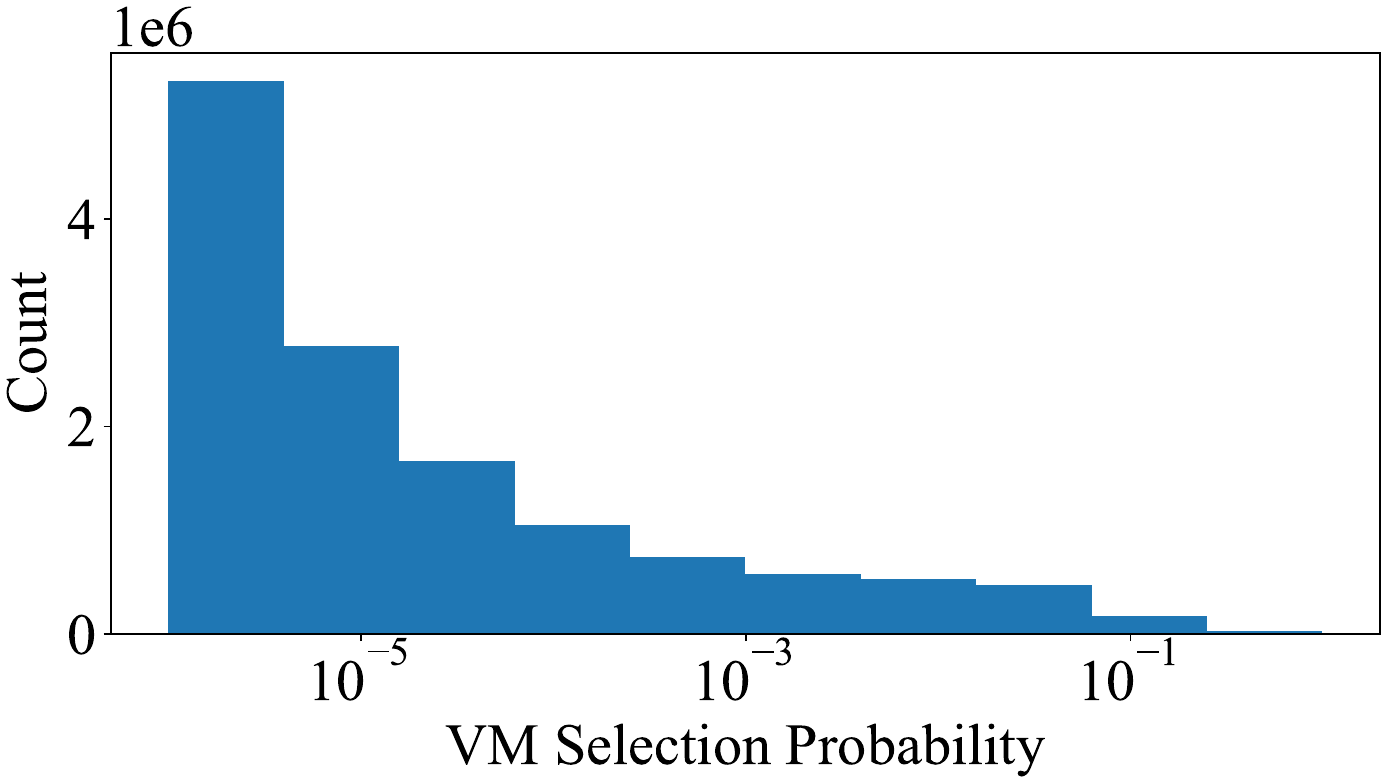}
            \vspace{-0.1in}
		\caption{VMs probability distribution.}
		\label{fig: ablation-distribution}
	\end{minipage}
		\begin{minipage}[t]{0.32\linewidth}
		\centering
		 \includegraphics[width=2.2in,height=1.2in,angle=0]{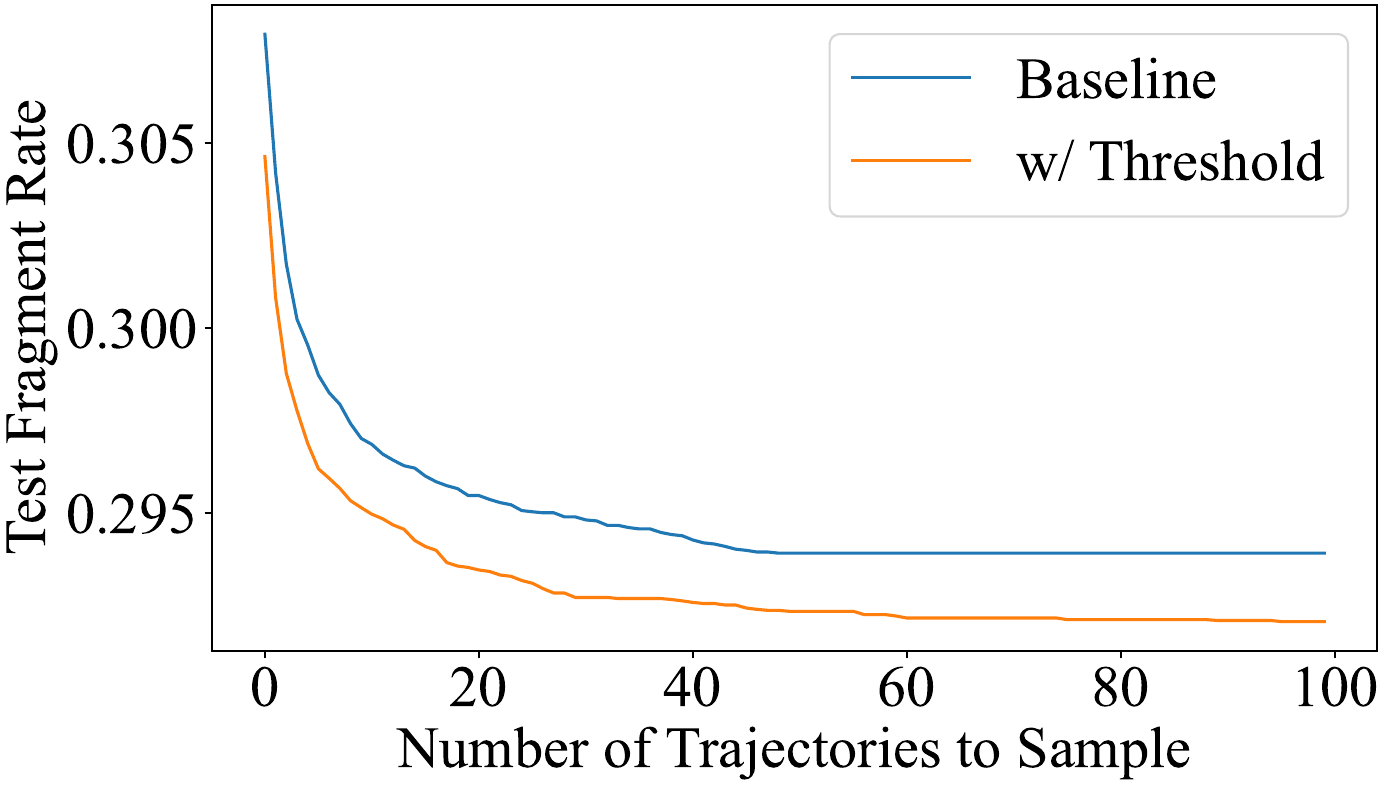}
            \vspace{-0.1in}
		\caption{Risk-seeking Evaluation.}
		\label{fig: ablation-risk-seeking}
	\end{minipage}
	\vspace{-0.1in}
\end{figure*}

\noindent \textbf{Monte-Carlo Tree Search (MCTS) \cite{Zhu21CIKM}:} As traditional search based methods need to perform multiple rollouts during inference time to achieve a good performance, we use DDTS \cite{Zhu21CIKM} to prune the search space.

\noindent \textbf{Decima \cite{mao2019learning}:} Decima uses a graph neural network to encode the VM and PM information and trains using deep RL. Decima balances the size of the action space and the number of actions required by decomposing VM rescheduling decisions into a two-dimensional action, which outputs i) the VM that needs to migrate, and ii) an upper number of PM subsets to choose as the destination. 

\noindent \textbf{NeuPlan \cite{zhu2021network}:} In the first stage, an RL agent takes in the problem as a graph and generates the first few VM migrations to prune the search space. In the second stage, it uses an MIP solver for the remaining MNLs. A relax factor $\beta$ (30 in our case) is used to control the size of the MNL space to be explored by MIP.

\subsection{Overall Performance}\label{sec: fr_time_performance}
Fig. \ref{fig: motivation_exp_rl} shows the FR and inference latency of all methods on the Medium dataset.
\aliasAPP achieves a lower FR compared to all baselines. Notably, \aliasAPP is merely \textbf{2.86\%} behind the optimal MIP solution ($0.2941$ vs. $0.2859$) when $MNL = 50$. Meanwhile, \aliasAPP can generate one trajectory within 1.1s, while MIP requires 50.55 minutes to provide the near-optimal solution. It is worth noting that with higher MNLs, the performance gap between \aliasAPP and MIP does not increase as significantly as compared to other baselines.

$\alpha$-VBPP only removes $\alpha$ number of the worst VMs for
each stage based on a single timestep, failing to consider
future opportunities to replace them back, which leads to its
inferior performance. POP fails to achieve good performance since it still relies on MIPs to solve each subproblem. To meet the second-level latency requirement, we must divide the problem into many subproblems, causing its solutions to be only locally optimal. On the other hand, Decima reduces the large action space by limiting the PM actor to only select from a subset of PMs, but the subsampling of PMs is completely random, as opposed to our solution. While MCTS with DDTS uses neural networks to prune the search space, it still requires a significant number of rollouts to achieve stable performance. Lastly, NeuPlan also fails to deliver a satisfying solution for high MNLs, since it can only use MIP to solve a small number of steps in order to meet the latency requirement. Although HA/MCTS achieves competitive results on smaller MNLs (MNL $\leq$ 20), repeating HA/MCTS with smaller MNLs multiple times, as done by $\alpha$-VBPP, performs poorly on larger MNLs (e.g., MNL=50) because it tends to get stuck in local optima at each individual stage.

To summarize, algorithms that involve MIP or searching often fail to deliver a satisfying solution under the strict latency requirement. Heuristic methods are fast but are also suboptimal. Deep learning-based methods can meet the latency requirement since the models do not need to be retrained at inference time, but are difficult to train without the set of customized techniques we proposed for VMR. We provide a case study and a tool to visualize where the improvements of \aliasAPP come from in Section \ref{sec: case-study}.

\subsection{Performance Decomposition of \aliasAPP}\label{sec: component}

We conduct an ablation study using MNL = 50 on the Medium dataset. In summary, FR performance reduces 16.46\% without the two-stage framework\footnote{The results for the two-stage framework are shown in Section \ref{sec: diff-constraint}.}, and improves from 0.3090 and 0.3079 to 0.2941 when sparse attention and risk-seeking are added, respectively. Recall that the near-optimal MIP solution is 0.2859, which means that \textit{Sparse Attention} achieves $\frac{0.3090 - 0.2941}{0.3090 - 0.2859} = 64.5\%$, and \textit{Risk-Seeking} achieves $\frac{0.3079 - 0.2941}{0.3079 - 0.2859} = 62.7\%$ of the potential room for improvement.

\noindent \textbf{Sparse Attention.}
We compare against two baselines. \textit{w/o Attention} uses a multilayer perceptron (MLP)~\cite{popescu2009multilayer} as the feature extraction module. MLP concatenates features of all PMs and VMs and thus requires much more trainable parameters that scale linearly with the number of machines in the system. \textit{Vanilla Attention} has fewer parameters as it shares a single small embedding network for all PMs and a second small embedding network for all VMs, but it uses the original encoder-decoder transformer~\cite{vaswani2017attention, yang2024orchloc} without attending to tree-level features. Fig. \ref{fig: ablation-sparse-attn} shows that MLP fails to converge due to its large number of trainable parameters. As training progresses, \textit{Sparse Attention} learns to capture relational features unique to VMR and gradually outperforms \textit{Vanilla Attention}. We show a case study of how such relational information can benefit VMR intuitively in Section \ref{sec: case-study}.

\noindent \textbf{Risk-Seeking Evaluation.}
At deployment time, given a trained \aliasAPP checkpoint and an initial vm-pm state, we generate multiple migration plans. We then leverage our simulator to calculate the resulting objective under each plan and only deploy the plan that yields the best outcome. Recall that \aliasAPP takes 1.1s to generate each trajectory, and suppose we have 8 GPUs to run generations in parallel, generating 16 trajectories would take 2.2s.

To avoid sampling suboptimal actions in the trajectory, we mask out PMs and VMs that are assigned low probabilities. We plot the distribution of probabilities assigned to different VMs in the validation set in Fig. \ref{fig: ablation-distribution}. Notice that most VMs are assigned low probabilities. In fact, fewer than $0.8\%$ of VMs have a greater than $1\%$ chance of being selected. Therefore, we compute a quantile $\in \{0.95, 0.98, 0.99, 0.995\}$ for all VMs and all PMs at each step and mask out all machines that have probabilities that fall below the corresponding threshold. We perform a grid search on the two quantiles using the validation set and apply the best combination to the test set. Fig. \ref{fig: ablation-risk-seeking} shows the final FR decreases when we sample more trajectories, and the FR decreases further after we apply thresholding.

\begin{figure}[t]
\begin{center}
  \includegraphics[height=1.5in, width=3.2in]{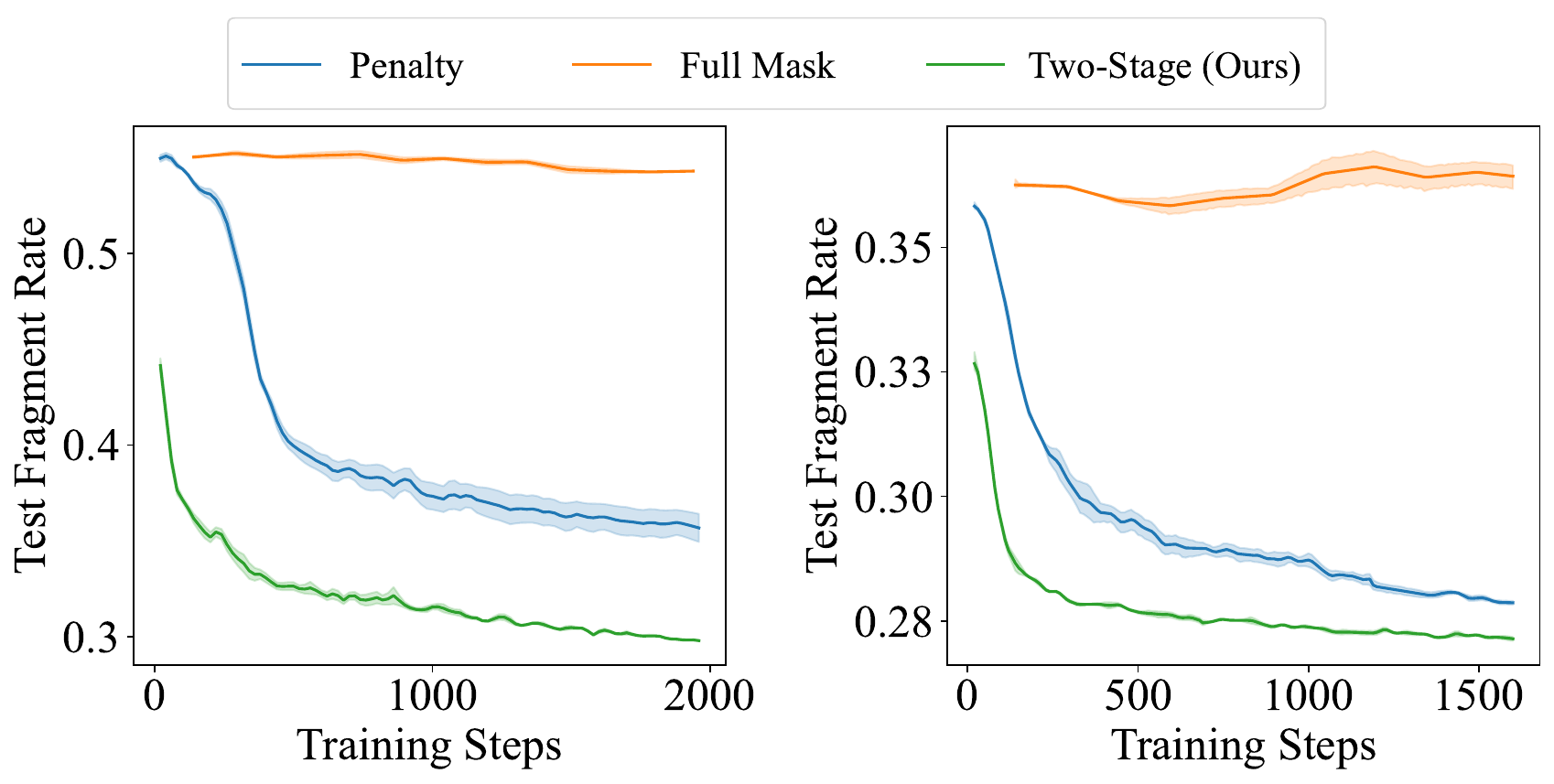}
  \end{center}
    \vspace{-0.2in}
  \caption{Constraints on Medium (left) and Multi-Resource (right).}
  \label{fig: constraint-multi}
  
\end{figure}

\begin{figure}[t]
\begin{center}
  \includegraphics[height=2.0in, width=2.8in]{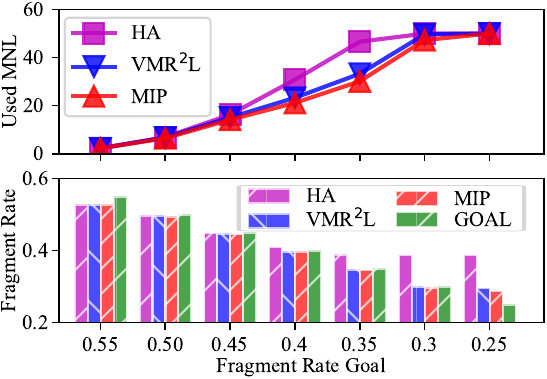}
  \end{center}
  \vspace{-0.2in}
  \caption{MNL performance under different FR goals.}
  \label{fig: minimize_mnl}
\end{figure}

 \begin{figure*}[t]
	\hspace{1ex}
		\begin{minipage}[t]{0.30\linewidth}
		\centering
		 \includegraphics[width=2.0in,height=1.0in,angle=0]{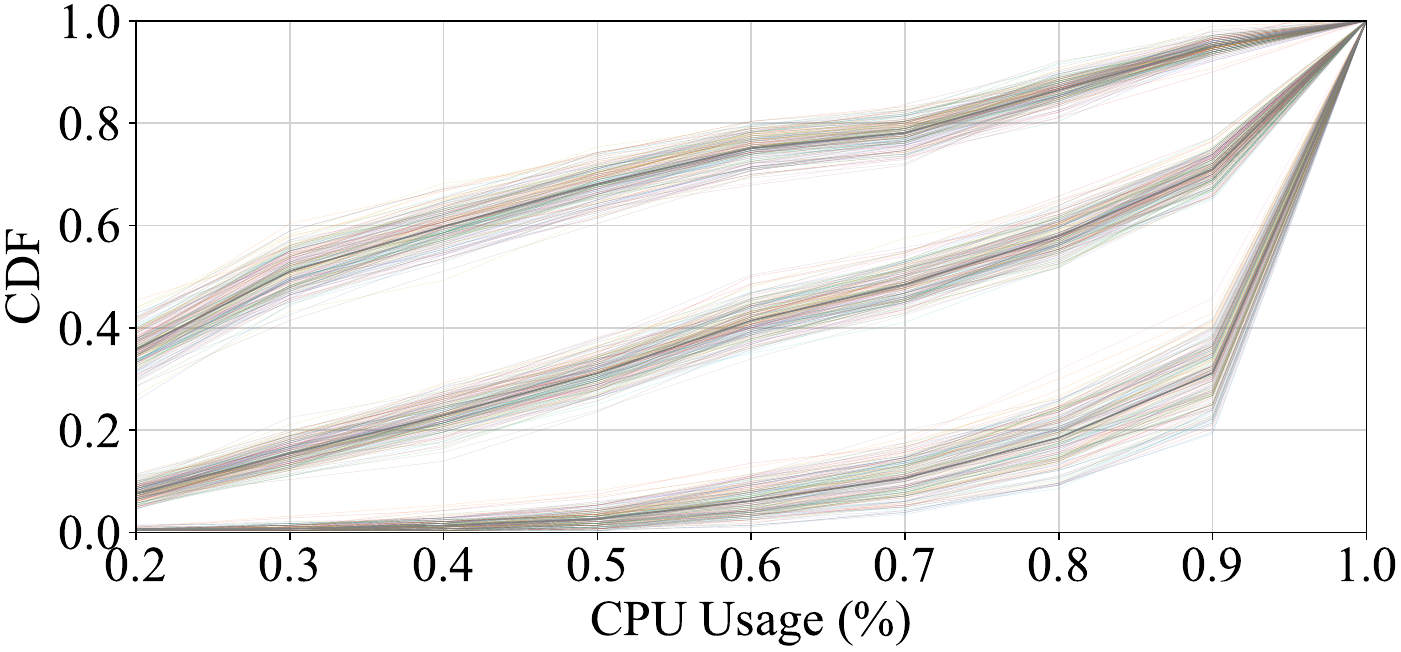}

		\caption{CPU usage on PMs under different workloads.}
		\label{fig: cpu_usage}
	\end{minipage}
		\begin{minipage}[t]{0.30\linewidth}
		\centering
		 \includegraphics[width=2.0in,height=1.0in,angle=0]{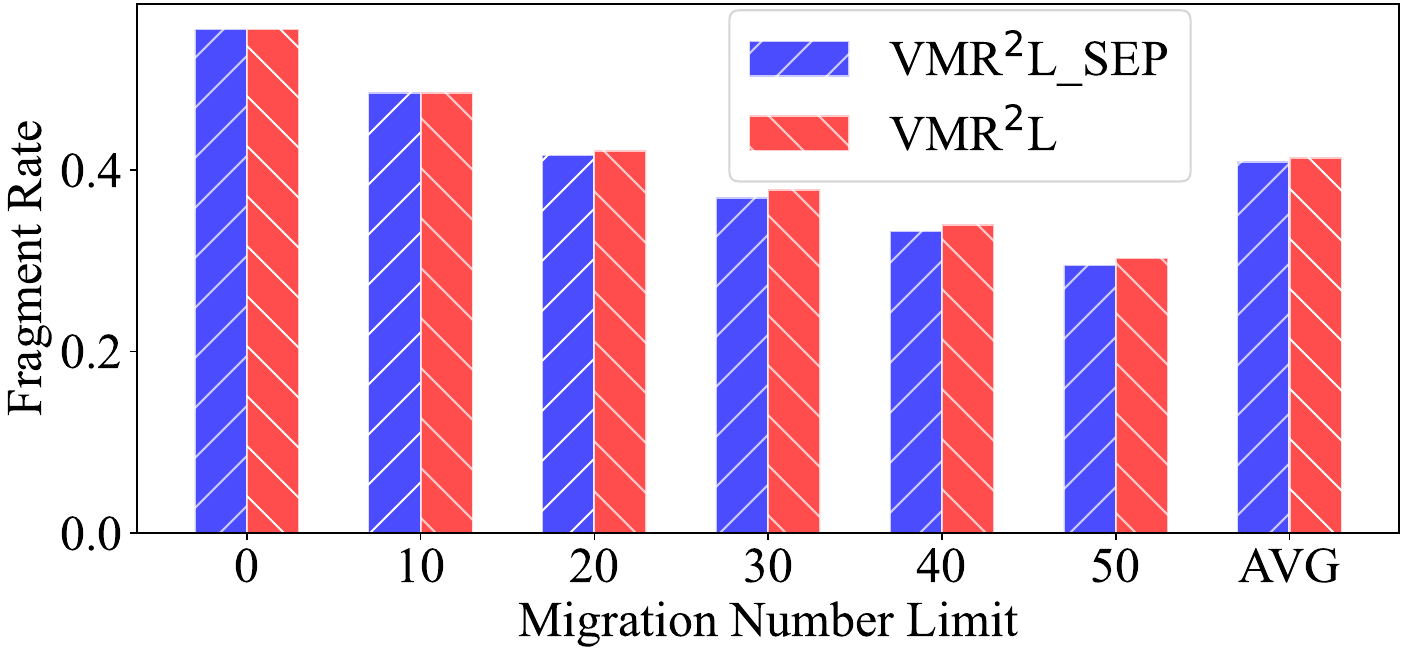}
		\caption{FR gap between \aliasAPP and \aliasAPP$_{\text{SEP}}$.}
		\label{fig: fr_separate}
	\end{minipage}
		\begin{minipage}[t]{0.36\linewidth}
		\centering
		 \includegraphics[width=2.4in,height=1.1in,angle=0]{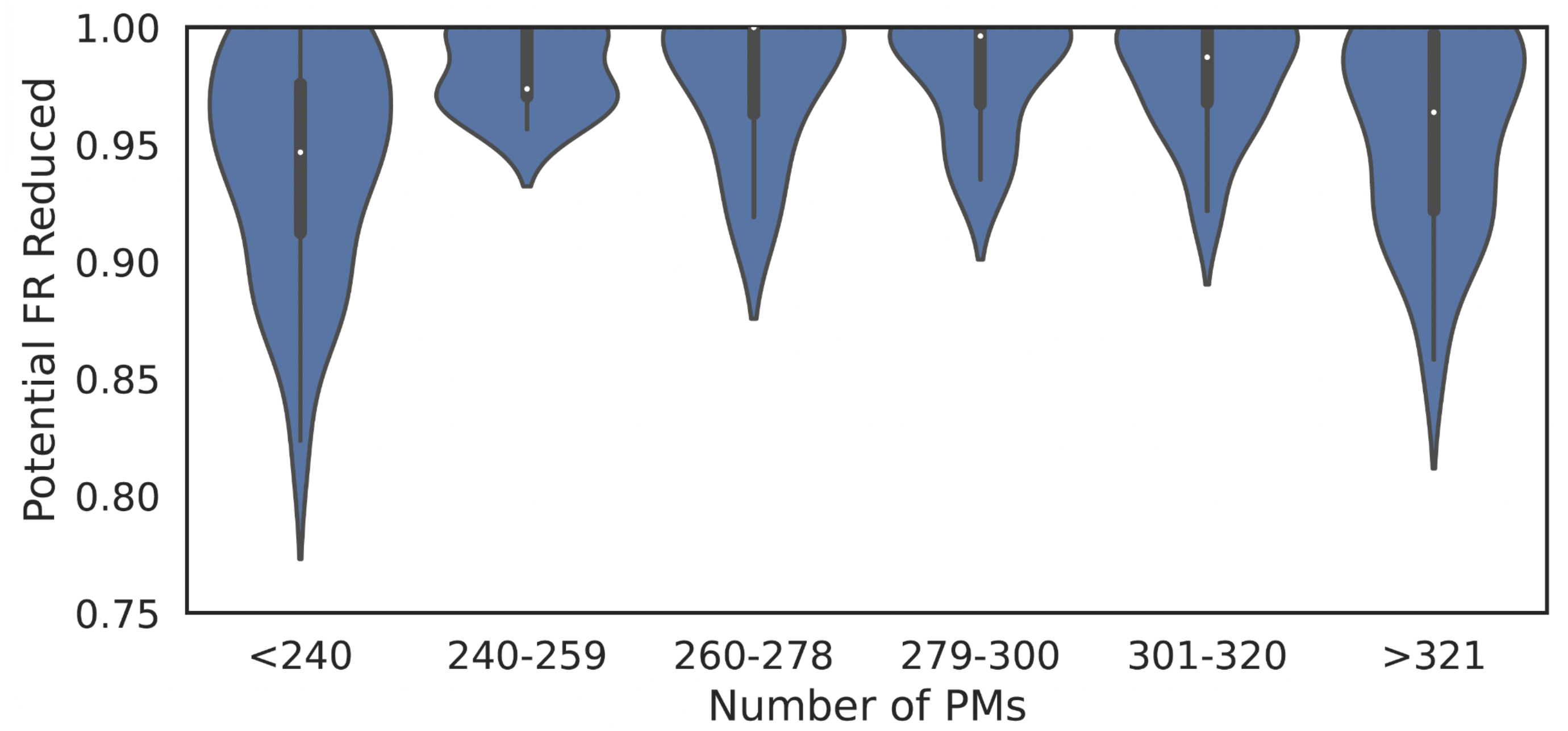}

		\caption{Ratio of potential FR achieved when deploying on clusters with different numbers of PMs.}
		\label{fig: diff-clusters}
	\end{minipage}
 
	% \vspace{-0.1in}
\end{figure*}

\begin{table}
\caption{FR under different affinity constraint levels.}
  \label{tab: affinity}
  \footnotesize
  \centering
\begin{tabular}{l||llllll}

    \hline
Aff. Level      & 0      & 1      & 2      & 3      & 4      & 8       \\
    \hline
Aff. Ratio & 0      & 1.12\% & 1.86\% & 3.46\% & 6.50\% & 38.3\% \\
\aliasAPP                  & 0.3032 & 0.3029  & 0.3034 & 0.3048  & 0.3045  & 0.3306  \\
% Avg anti-affinity   & 0      & 23     & 39     & 72     & 136    & 800     \\
MIP                 & 0.2859 & 0.2860 & 0.2860 & 0.2862 & 0.2864 & OOT  \\  

    \hline
\end{tabular}

\end{table}

\subsection{Different Constraints with Two-Stage Framework}\label{sec: diff-constraint}

\noindent \textbf{More Resource Constraints.} To analyze how the two-stage framework supports different constraints, we compare it with two baselines: \textbf{i) Penalty}: a penalty of \(-5\) is given if the agent takes an illegal action, and \textbf{ii) Full-Mask}: the VM candidate and the PM destination are generated simultaneously, with all illegal pairs having probabilities of zero.

In addition to the Medium dataset, we consider traces from an additional data center that is smaller but has more complicated multi-dimensional resource constraints with more VM and PM types. The new \textit{Multi-Resource} dataset has two PM types --- one has 88 CPUs and 256 GB of memory, and the other has 128 CPUs and 364 GB of memory. The regular VM types in Table \ref{tab: vmtype} always have a CPU-memory ratio of $1:2$. For memory-intensive applications, a user might request additional memory, and the CPU-memory ratio can increase up to $1:8$.

As we can see from Fig. \ref{fig: constraint-multi}, \textbf{Penalty} suffers from a slower convergence to a sub-optimal level, since the large negative penalties required to eliminate illegal actions tend to dominate the gradient signal, especially during early stages of training. \textbf{Full-Mask} does not converge under both constraint settings, since its action space is the product of the number of VMs and the number of PMs, which leads to poor exploration. In comparison, \textbf{Two-Stage} decomposes the action space by designating stage 1 to focus on the set of VM candidates and stage 2 to focus on the set of PM destinations, resulting in much faster convergence. 

\noindent \textbf{Service Constraints.} We consider a practical constraint in the form of a hard anti-affinity, where a VM cannot be placed on the same PM with some other selected VMs. It can i) prevent resource-intensive VMs to be hosted together, which leads to performance interference, and ii) support critical services that require backup VMs in case of a PM failure. To enforce anti-affinity, we mask out all PMs that host conflicting VMs in stage 2 after selecting a VM candidate in stage 1. We typically observe an affinity ratio requirement to be under $5\%$ in real-world traces. Affinity ratio means the average percentage of VMs that a given VM conflicts with. We synthesize the service anti-affinity constraint on the Medium dataset.

In Table \ref{tab: affinity}, we see that \aliasAPP maintains consistent performance under typical affinity ratio levels. To demonstrate the robustness of \aliasAPP to extreme constraint levels, we further evaluate it when the affinity ratio surges to $38.3\%$ and see that \aliasAPP is still able to achieve a reasonable FR.

\subsection{Objectives Generalization}\label{sec: diff-obj}
\subsubsection{Minimize MNL Given FR Goals.}
\aliasAPP's flexibility enables it to learn different policies depending on the high-level objective. We now consider a new objective: 
%\deleted{we would like to }
minimize the number of VM migrations to reach a given FR goal, to reduce migration costs. To support this objective, we simply modify the original reward function (Equation \ref{eq: original-reward}) as follows:
\begin{align}\label{eq: new_reward}
R_{fr} &= (S_{\text{before, src}} - S_{\text{after, src}}) + (S_{\text{before, dest}} - S_{\text{after, dest}}), \\
R &= \left\{ \begin{array}{lr}
                  -1 + R_{fr} ,  &FR > FR_{\text{Goal}},\\
                 10 + R_{fr} ,  &FR \leq FR_{\text{Goal}}.  \\
            \end{array}
    \right.
\end{align}

On top of the original reward, we add a penalty of -1 if the FR is above the goal as it indicates additional VM migrations are required, and a bonus of +10 if \aliasAPP reaches the goal. In Fig. \ref{fig: minimize_mnl}, the top subfigure shows the number of migration steps, while the bottom subfigure shows the achieved FR, both sharing the x-axis as different FR goals.
On average, MIP and \aliasAPP achieve 14.77\% and 11.11\% fewer MNLs than HA, respectively. \aliasAPP requires only 3.66\% more VM migrations, but with second-level solution time.

\subsubsection{Mixed Objective (i): Multi-VM-Type FR}
What if a cluster wants to optimize for multiple VM types, such as 16xlarge VMs in addition to the original 4xlarge VMs? Recall that 4xlarge requires 16 cores on a single NUMA, while 16xlarge requires 64 cores deployed across two NUMAs, introducing additional complexities. Let $FR_{16}$ denote 16-core FR, and $FR_{64}$ denote 64-core FR. We optimize for the objectives as the convex combinations of $FR_{16}$ and $FR_{64}$:

\begin{equation}\label{eq: mix-obj}
    \text{Obj}_\lambda = \lambda \cdot FR_{64}  +  (1 - \lambda) \cdot FR_{16} ,
\end{equation}

where $\lambda$ is a predefined parameter based on which VM types to prioritize. Table \ref{tab: mixed-obj-one} presents the results when we optimize for $\lambda \in \{0, 0.2, 0.4, 0.6, 0.8, 1\}$ on the \textit{Multi-Resource} dataset introduced in Section \ref{sec: diff-constraint}. For both methods, as $\lambda$ increases, some $FR_{16}$ has to be sacrificed when the objective emphasizes more on $FR_{64}$. Overall, in terms of Obj$_\lambda$, we see that \aliasAPP consistently outperforms POP under different $\lambda$'s. 
Note that when $\lambda = 0.2$, Obj$_\lambda$ is lower for \aliasAPP, but POP achieves lower $FR_{64}$. This is because to minimize a mixed objective, different algorithms may choose to focus on each individual objective differently, even under the same $\lambda$. Therefore, individual objectives might not be directly comparable.

\begin{table}
  \caption{Mixed objective (i) with $FR_{16}$ and $FR_{64}$.}
  \label{tab: mixed-obj-one}
  \footnotesize
  \centering
  \begin{tabular}{l|l||l|l|l|l|l|l}
    \hline
      &\textbf{$\lambda$}& 0 & 0.2 & 0.4 & 0.6 & 0.8 & 1\\
    \hline
    \textbf{\aliasAPP} & \textbf{$FR_{16}$} & 0.2941 & 0.3079 & 0.3413 & 0.4086 & 0.4214 & 0.4532 \\       
    & \textbf{$FR_{64}$} & 0.9478 & 0.9263 & 0.6960 & 0.5900 & 0.5603 & 0.5473 \\
    \hline
    & \textbf{Obj}$_\lambda$ & 0.2941 & 0.4316 & 0.4832 & 0.5174 & 0.5325 & 0.5473 \\
    \hline
    \hline
    \textbf{POP} &\textbf{$FR_{16}$} & 0.3447 & 0.3650 & 0.3971 & 0.3992 & 0.3991 & 0.5215 \\       
    &\textbf{$FR_{64}$} & 0.9836 & 0.8235 & 0.7312 & 0.7280 & 0.7278 & 0.7222 \\ 
    \hline
    & \textbf{Obj}$_\lambda$ & 0.3447 & 0.4567 & 0.5308 & 0.5964 & 0.6621 & 0.7222 \\
    \hline
  \end{tabular}
\end{table}

\subsubsection{Mixed Objective (ii): Multi-Resource-Type FR}
In some clusters, the optimization target can be defined in terms of multiple resource types. We show that \aliasAPP is capable of handling such a multi-dimensional objective by using 16-core CPU fragments and 64-GB Memory fragments (denoted as $Mem_{64}$) as an example. $Mem_{64}$ refers to a discrete resource fragment representing 64 GB of memory capacity. The objective function can be written in a form similar to Equation \ref{eq: mix-obj}. We show the results on the \textit{Multi-Resource} dataset in Table \ref{tab: mixed-obj-two}. \aliasAPP still consistently outperforms POP in terms of the mixed objective. Another interesting observation is when we increase $\lambda$ from 0 to 0.2, both $FR_{16}$ and $Mem_{64}$ decreased for \aliasAPP. We hypothesize that rewarding the model for additional objectives helps to make the reward less sparse, thereby further stabilizing RL training \cite{pmlr-v80-riedmiller18a}.

\begin{table}
  \caption{Mixed objective (ii) with $FR_{16}$ and $Mem_{64}$.}
  \label{tab: mixed-obj-two}
  \footnotesize
  \centering
  \begin{tabular}{l|l||l|l|l|l|l|l}
    \hline
      &\textbf{$\lambda$}& 0 & 0.2 & 0.4 & 0.6 & 0.8 & 1\\
    \hline
    \textbf{\aliasAPP} & \textbf{$FR_{16}$} & 0.2709 & 0.2700 & 0.2872 & 0.3071 & 0.3127 & 0.3182 \\       
    & \textbf{$Mem_{64}$} & 0.3032 & 0.2955 & 0.2695 & 0.2480 & 0.2490 & 0.2449 \\
    \hline
    & \textbf{Obj}$_\lambda$ & 0.2709 & 0.2751 & 0.2802 & 0.2716 & 0.2617 & 0.2449 \\
    \hline
    \hline
    \textbf{POP} &\textbf{$FR_{16}$} & 0.2808 & 0.2809 & 0.2843 & 0.3055 & 0.3073 & 0.4464 \\       
    &\textbf{$Mem_{64}$} & 0.3107 & 0.2832 & 0.2762 & 0.2559 & 0.2550 & 0.2537 \\ 
    \hline
    & \textbf{Obj}$_\lambda$ & 0.2808 & 0.2813 & 0.2811 & 0.2757 & 0.2655 & 0.2537 \\
    \hline
  \end{tabular}
\end{table}

\subsection{Generalization and Scalability of \aliasAPP}\label{sec: generalization}

\begin{table}
  % \vspace{-0.2in}
  \caption{Generalization to abnormal workloads. POP is a baseline algorithm for large-scale resource allocation \cite{narayanan2021solving}.}
  \label{tab: workloads}
  \footnotesize
  \centering
  \begin{tabular}{l||l|l|l}
    \hline
     Methods& L (MNL=100)  & M (MNL=100)& H (MNL=50)      \\

    \hline
    HA & 0.256(-2.7\%) & 0.276(-8.0\%) & 0.387(-10.9\%)\\       
     \aliasAPP (L)  & \textbf{0.237(+4.8\%)} & 0.261(-2.7\%) & 0.424(-18.6\%)\\
     \aliasAPP (M)  & 0.239(+4.0\%) & 0.238(+6.3\%) & 0.422(-18.2\%)\\
     \aliasAPP (H)  & 0.243(+2.4\%) & 0.248(+2.4\%) & \textbf{0.303(+12.2\%)}\\
     \aliasAPP (L,H) & \textbf{0.237(+4.8\%)} & \textbf{0.237(+6.7\%)} & 0.326(+5.5\%)\\
    \hline       
    POP & 0.249 & 0.254 & 0.345\\
    \hline
  \end{tabular}

\end{table}

\subsubsection{Abnormal Workloads}\label{sec: abnormal_workloads}
It is well-known that common DRL applications often experience performance degradation due to distribution shifts \cite{Levine2020OfflineRL}. However, in real-world service traces, there are periods, such as during deadlines or holidays, where workloads (defined as the percentage of available CPUs on PMs) deviate significantly from the norm. To assess whether \aliasAPP can adapt to these abnormal workload levels, we gathered two additional datasets representing \textit{Low}(L) and \textit{Middle}(M) workloads. In our study, the medium dataset represents \textit{High}(H) workloads. The workload distributions of each train mapping from the three datasets are shown in Fig. \ref{fig: cpu_usage}. Note that these three datasets have strictly non-overlapping workload distributions, i.e., we cannot find a training sample from \textit{High} that has a workload similar to \textit{Middle} or \textit{Low}.

We train \aliasAPP on one or mixed workload datasets and evaluate the FR on each individual workload level. The results are summarized in Table \ref{tab: workloads}. For example, (L,H) means that we train \aliasAPP on both \textit{Low} and \textit{High} datasets. We set $MNL=100$ for L and M since the FR difference is low until MNL increases to $100$. As MIP runs out of time due to the larger MNL used on L and M, we choose POP as the standard baseline since it is easy to tune and exhibits strong FR performances. First, we see that \aliasAPP outperforms the two baselines when trained on the same workload level as test. When \aliasAPP trains with workloads smaller than the test workload, \aliasAPP suffers from performance degradation. Intuitively, a high workload requires the agent to create available resources by first removing existing VMs away from the destination PM, but this action is less common on lower workloads and thus cannot be effectively learned. 

Remarkably, when trained on both L and H, \aliasAPP can learn a general policy and perform the best on M \textit{without ever experiencing middle workloads}. Thus, we recommend end-users to train \aliasAPP with a combination of high and low workloads from their data centers. This allows \aliasAPP to bridge gaps in the training data and better generalize when encountering abnormal workloads.

\subsubsection{Generalizing to Different MNLs}\label{sec: exp-generalize}
In practical scenarios, MNL often fluctuates due to varying business needs, such as when in the day VMR is performed. We show that training a single \aliasAPP agent with $MNL=50$ can yield effective results across a range of MNLs $\in \{10, 20, 30, 40, 50\}$. To compare, we train a separate \aliasAPP agent for each MNL, denoted as \aliasAPP$_{\text{SEP}}$. As shown in Fig. \ref{fig: fr_separate}, \aliasAPP performs only marginally worse than \aliasAPP$_{\text{SEP}}$ with an average FR performance gap of 1.16\%. This suggests that the \aliasAPP agent trained with a large MNL can be readily applied to tasks with smaller MNLs. It avoids the overhead of maintaining a separate \aliasAPP agent for each MNL.

\subsubsection{Generalizing to Different Clusters}\label{sec: generalization_pms}
We evaluate the generalization ability of the \aliasAPP when trained on a specific cluster and deployed to different clusters. Specifically, we use the \aliasAPP model trained on the Medium dataset, which contains 280 PMs. To simulate varying cluster sizes, we randomly modify the Medium dataset by adding or removing PMs, generating a total of 100 different mappings.

\begin{figure}[t]
\centering
\includegraphics[width=2.5in,height=1.2in]{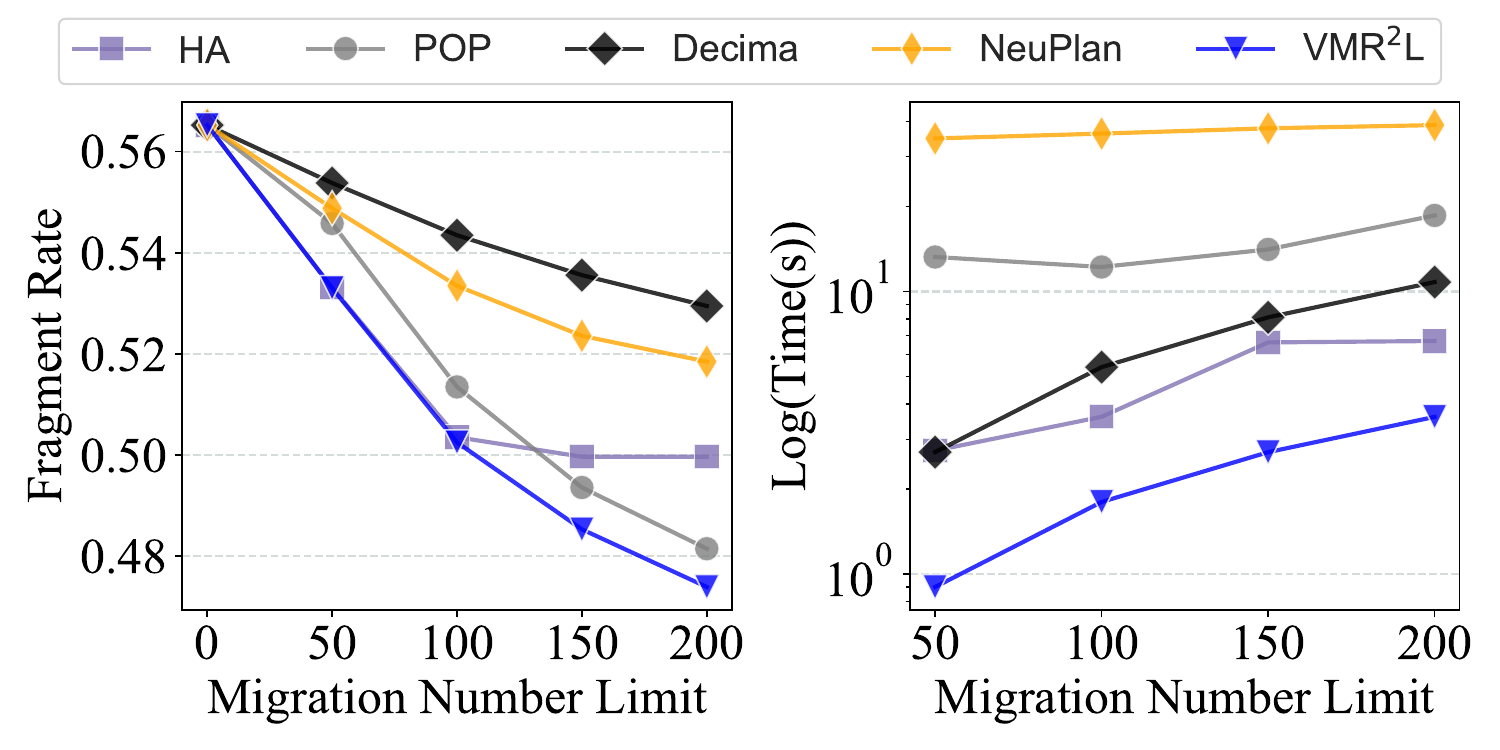}
\caption{FR and time performance on the Large dataset.}
\label{fig: fr_time_large_datasaet}
\end{figure}

Fig. \ref{fig: diff-clusters} illustrates the potential FR achieved by \aliasAPP on clusters with different numbers of PMs. The potential FR is defined as the difference between the initial FR and the FR achieved by MIP. Our results show that \aliasAPP maintains nearly the same performance when deployed on clusters with a PM count that varies by less than $10\%$. Performance remains robust even with variations of $10\%$ to $20\%$, achieving over $95\%$ of the potential FR. However, when the number of PMs differs by more than $20\%$, a slight decline in performance is observed.  Even in this scenario, \aliasAPP still significantly outperforms POP, which achieves only around $78\%$ and has to be retrained on each cluster.

\subsubsection{More VMs \& PMs.} To see how \aliasAPP scales to a larger cluster, we conduct experiments on the Large dataset with 4546 VMs and 1176 PMs. Fig. \ref{fig: fr_time_large_datasaet} shows the performance against different baselines when MNL varies from 50 to 200. MIP is not included here since it takes more than an hour to solve a single migration path. Similar to the Medium dataset, \aliasAPP achieves lower FRs than the baselines, with an inference time of 3.8s to solve one mapping.

\begin{figure}[t]
\centering
\subfigure[FR on the low workload.]{
\includegraphics[width=1.5in,height=1.4in]{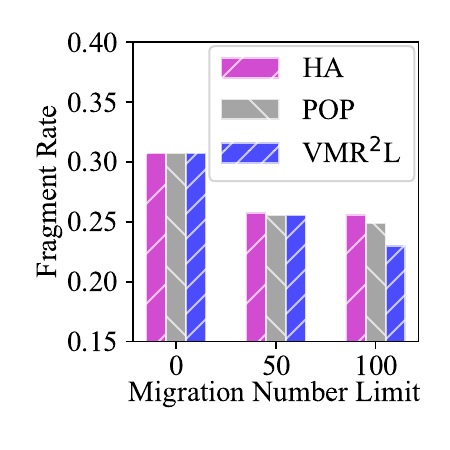}
\label{fig: small_workload_fr}
}
%\quad
\subfigure[FR on the medium workload.]{
\includegraphics[width=1.5in,height=1.4in]{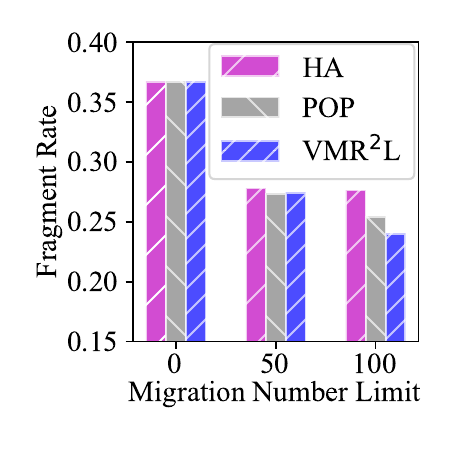}
\label{fig: middle_worklod_fr}
}
\vspace{-0.2in}
\caption{FR on different workloads.}
\label{fig: different_workload}

\end{figure}

\subsubsection{Different Workloads with Different MNLs}\label{sec: workload_mnl}
We evaluate \aliasAPP with varying workloads under different MNLs. We set MNL=100 when evaluating the Low and Middle workload datasets since the FR difference is low until we increase MNL to 100. From Fig. \ref{fig: different_workload}, we can see HA, POP, and \aliasAPP can all decrease FR at MNL=50. However, HA fails to decrease FR at MNL=100. Instead, \aliasAPP achieves 7.42\% and 4.8\% lower FR on the low workload, and 13.77\% and 6.3\% lower FR on the middle workload compared to HA and POP, respectively. These results demonstrate that \aliasAPP is capable of handling varying workloads.

\subsection{Is A Larger Cluster More Difficult for \aliasAPP to Learn?}\label{sec: larger_cluster}

\begin{figure}[t]
\centering
\subfigure[Including the initial stage.]{
\includegraphics[width=2.7in,height=1.6in]{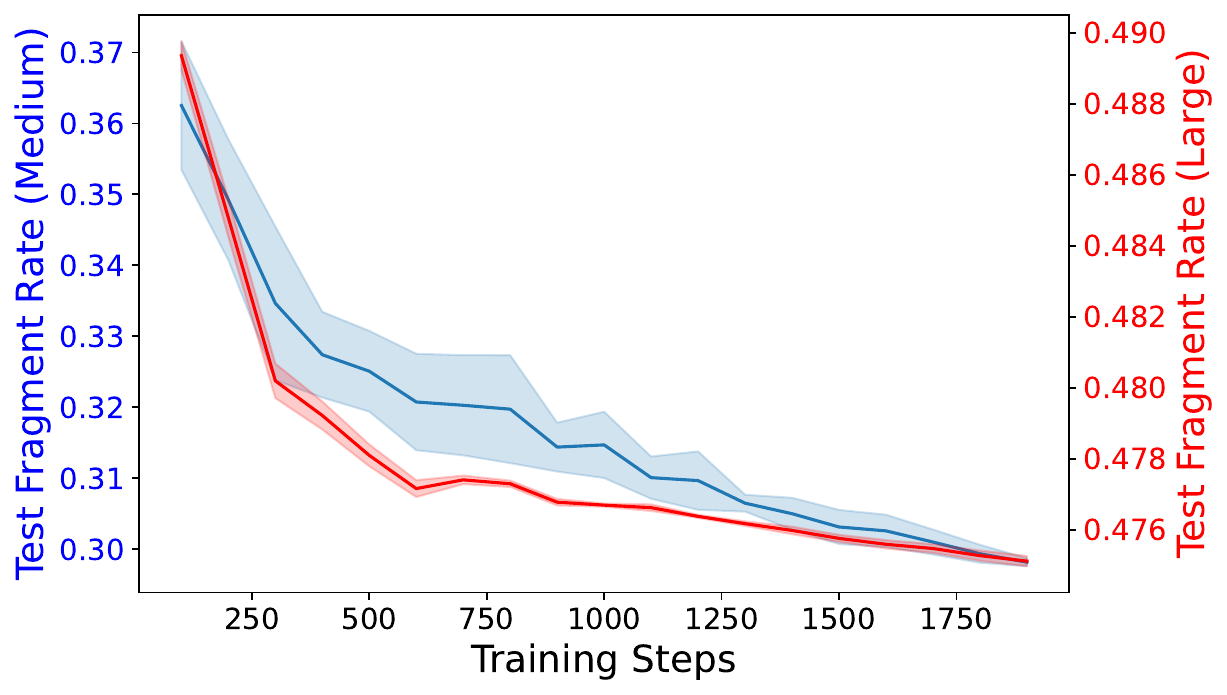}
\label{fig: large-converge-before}
}
\vspace{-0.1in} % Adjust vertical spacing between the subfigures
\subfigure[After the initial stage.]{
\includegraphics[width=2.7in,height=1.6in]{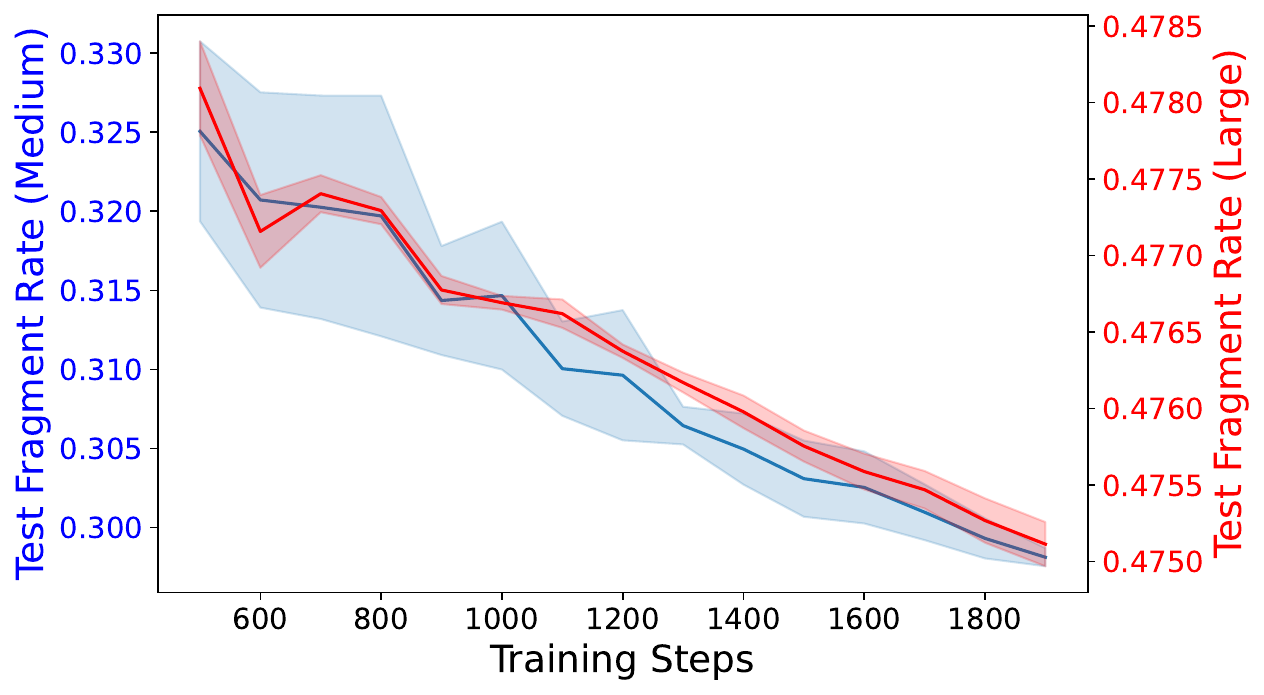}
\label{fig: large-converge-after}
}
\vspace{-0.1in}
\caption{Convergence speed on different cluster sizes.}
\label{fig: large-converge}
\end{figure}

Recall that the Medium dataset has up to 2089 VMs and 280 PMs, and the Large dataset has up to 4546 VMs and 1176 PMs. We demonstrate that larger clusters are not inherently more difficult to train in \aliasAPP by comparing the convergence speeds on the Medium and Large datasets. The results are shown in Fig. \ref{fig: large-converge-before}. Since the initial and optimal FR values are different between the two datasets, we use a dual y-axis to plot the convergence curves. At first glance, it may seem that convergence is slower on the Medium dataset (blue) compared to the Large dataset (red), which might seem counterintuitive given the higher number of VMs and PMs in the latter. However, we hypothesize that this is because there are smaller VMs in the Large dataset. Since smaller VMs are easier to move around, it is easier to learn a simple policy that can effectively reduce these fragments in the initial stages. To test this, we replot the convergence curves in Fig. \ref{fig: large-converge-after}, excluding the initial stage where most of the \quotes{low-hanging fruits} have been picked. After the initial stage, \aliasAPP converges slightly faster on the medium dataset, but the difference is very minimal and should not pose an issue. Note that since we use a dual y-axis, the absolute slopes are no longer comparable, but instead we shall focus on the similar linear downward trend in both datasets.

\begin{figure}[t]
\begin{center}
\includegraphics[height=2.8in, width=2.0in]{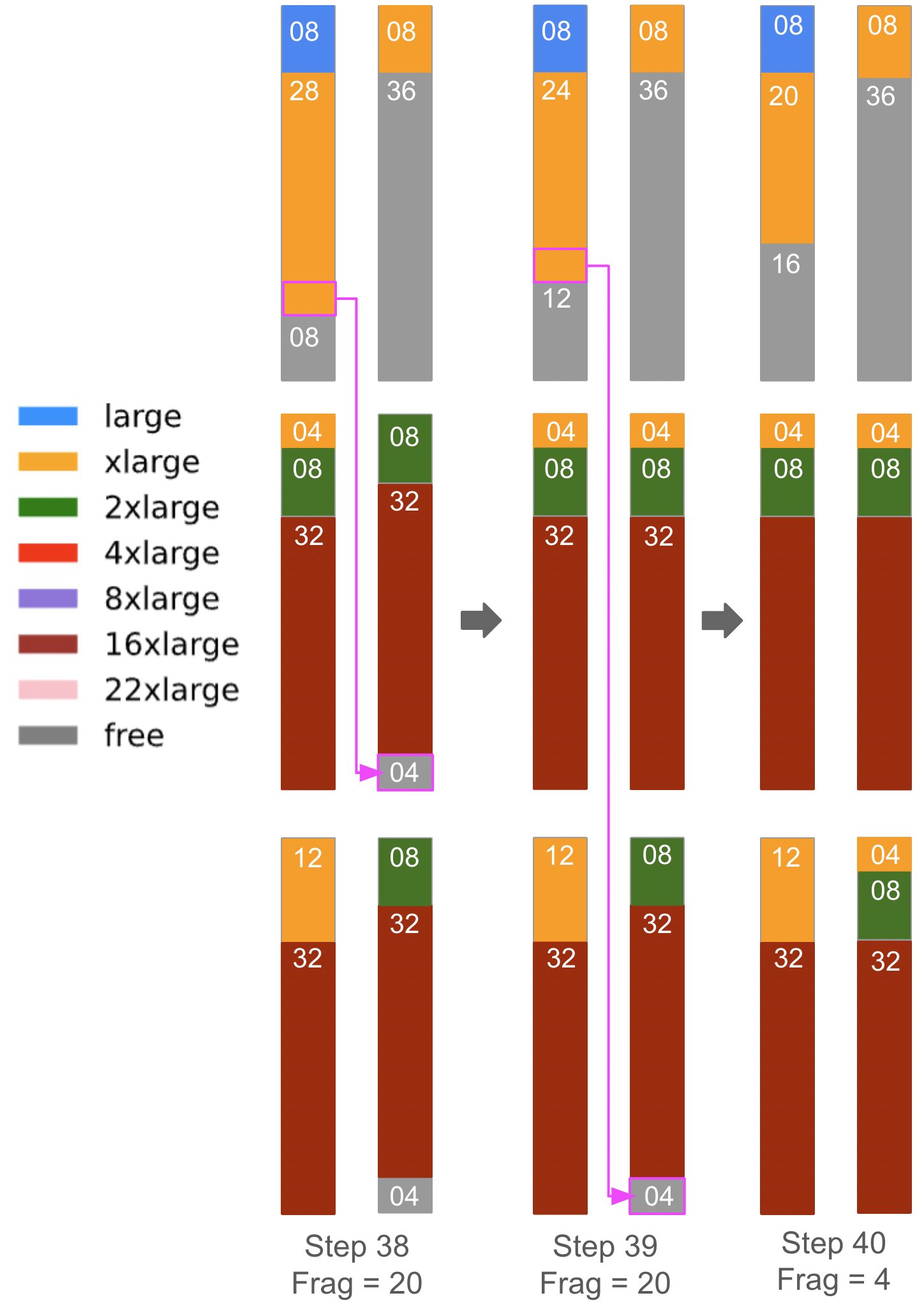}\\
\end{center}
% \vspace{-0.1in}
\caption{VM-PM Migration Details. Each equal-sized rectangle represents a NUMA node within a PM. Different colors indicate the total allocated size of a VM type on each NUMA.}
% \vspace{-0.1in}
\label{fig: reschedule_detail}
\end{figure}

\subsection{Intuitions Behind \aliasAPP: A Case Study} \label{sec: case-study}
Intuitively, where do the improvements of \aliasAPP come from? To answer this question, we build a tool to visualize which VM is being migrated at each step. We randomly select one mapping from 200 test mappings on the Medium dataset and analyze how \aliasAPP reduces FR when $MNL=50$. \aliasAPP optimizes FR from a global perspective. In Fig. \ref{fig: reschedule_detail}, we analyze the three PMs involved during steps 38-40. Each PM consists of two NUMA nodes, represented by the two vertically stacked bars. Different colors indicate the total allocated size of each VM type on a NUMA. For example, in the first NUMA at step 38, the orange section has a total size of 28 cores. Since an xlarge VM requires 4 cores, this implies that there are seven xlarge VMs occupying that space. The gray regions represent unused resources (free space).

At step 38, \aliasAPP removes a 4-core VM (from the orange section) from the top PM, which eliminates fragmentation on the destination NUMA but temporarily creates four fragments on the source NUMA. This results in a net reward of zero at this step. At step 39, the agent identifies a second 4-core VM on the source NUMA and migrates it to another PM, effectively eliminating all remaining fragments on both source and destination NUMAs. The reason the color distribution appears to change after migration is that the total allocated size of VMs on a NUMA is updated after each migration. The individual VMs remain the same, but their placement alters the total distribution of resources on each NUMA. This process highlights how sparse attention enables the agent to recognize multiple rescheduling opportunities and make globally optimal decisions for reducing FR across the entire system. This example shows that \aliasAPP is able to sacrifice immediate rewards for long-term FR performance due to the cumulative reward design in RL.

\section{Related Work} \label{sec: related-work}

\noindent \textbf{Connections to Bin Packing.} The use of optimized placement mechanisms proved to be successful in a broad set of use cases, including production quality scenarios \cite{ahmad2015survey} as well as transportation logistics \cite{Cai19DRL4KDD, Hu17, Duan19AAMAS, xia2016large}.
A typical solution exploits heuristics based on bin packing \cite{panigrahy2011heuristics}. In fact, VM placement can be modeled as a bin-packing problem, where VMs and PMs are objects and bins, respectively. Bin packing typically involves packing a set of items into fixed-sized bins such that the number of bins required \cite{Cai19DRL4KDD} or the total surface area is minimized \cite{Hu17, Duan19AAMAS}. However, there are two notable differences. First, the problem of VM rescheduling concerns adjusting an initial assignment of VMs to PMs. On the other hand, rebalancing items already packed in bins has received little attention in the context of other bin-packing applications. A critical aspect of the initial assignment is the current VM affiliations, which existing bin-packing solutions often do not consider but we show is critical to VMR via \textit{tree-level features}. Second, the total number of VMs and PMs in a data center can easily go into the range of several thousand or more \cite{xia2016large} and is far more than the typical scale of bin packing problems, which typically involve no more than a few hundred items \cite{Li18KDD, Zhu21CIKM}.

\noindent\textbf{RL for Optimization Problems.}
RL has been recently introduced to solve optimization problems, e.g., building ML compilers and optimizing neural network architectures \cite{haj2020autophase}. 
In particular, RL is used to select branching variables or find cutting planes in the Branch-and-cut method \cite{gasse2019exact,etheve2020reinforcement,gupta2020hybrid}.
Besides, RL can also be applied to existing heuristics for MIPs to further increase the quality of solutions \cite{song2020general,barrett2020exploratory}. In fact, most state-of-the-art solutions for optimization problems often involve MIP or searching \cite{zhu2021network}, but these methods are not directly appropriate for the VM rescheduling task due to their poor computation complexity. Although they are designed to accelerate MIPs, as shown in Section \ref{sec: fr_time_performance} even a state-of-the-art technique such as POP \cite{narayanan2021solving} fails to deliver a satisfying solution within the second-level time limits of the VM rescheduling task.
While learning-based methods \cite{mao2019learning} can meet the latency requirement by leveraging their generalization ability at deployment to avoid retraining, they do not involve techniques that are tailored for VMR, which we propose in \aliasAPP.

\section{Discussion}\label{sec: discussion}
\textbf{Noisy Neighbors.} A challenge in rescheduling is performance interference caused by noisy neighbors, which are VMs that disproportionately consume shared resources, leading to degradation for other VMs on the same PM. Our approach can address this issue by incorporating multi-resource constraints or hard anti-affinity policies (Section \ref{sec: diff-constraint}), ensuring that certain VMs are not hosted on the same PM to avoid resource contention. However, these strategies require prior knowledge of the resource profiles of the VMs involved and the ability to isolate resources accordingly. Future work may consider resource reservation to ensure dedicated resources without interference. Additionally, developing predictive models for workload characterization can help anticipate resource demands and interference patterns of different VMs, enabling more dynamic workload management.

\textbf{Adapting to New data.} Section \ref{sec: generalization} shows VMR\(^2\)L has strong generalization ability, learning a policy that performs well across different workload levels not present in the training data without retraining or even finetuning. It also generalizes to different MNLs by training a single agent with a larger MNL and applying it across smaller MNLs, avoiding the need for separate agents (Section \ref{sec: exp-generalize}). If we see large distribution shifts or performance drops, VMR\(^2\)L readily supports off-the-shelf finetuning methods (e.g., top-layer finetuning \cite{howard-ruder-2018-universal}, adding adapters \cite{houlsby2019parameter}, LoRA \cite{hu2021lora}).

\textbf{Efficient Training in Deterministic Environments.} VM rescheduling differs from typical RL tasks \cite{ding2024optimizing, ding2024multi} that require large datasets due to stochastic transitions. In contrast, VM rescheduling involves deterministic transitions, where the outcome is fully predictable given a specific state and action. We only require the initial VM mappings for training, making it more data-efficient. In light of this, our work introduces a simulator that enables offline training, fully simulating the rescheduling environment and allowing agents to learn policies without requiring extensive real-world data.

\textbf{Broader Insights.} While immediate application is scheduling, many system problems share the same underlying principles (e.g., large-scale, no environmental uncertainties, and strict latency requirements). Our RL formulation, and techniques such as
action decomposition and risk-seeking evaluation, are transferable and could inspire other system applications facing similar challenges.

\section{Conclusion}
Compared to conventional bin-packing applications, VM rescheduling presents unique challenges due to the expanding size of data centers. It must handle a large volume of VM requests while meeting a second-level inference speed, given the dynamic nature of VM states. As such, we propose \aliasAPP, a deep RL approach designed specifically for VM rescheduling: i) a two-stage framework to seamlessly accommodate different service constraints, ii) a sparse attention module to better capture local VM-PM relations, and iii) risk-seeking evaluation to offer a better trade-off between speed and performance. Future work includes optimizing for best-case performance during training \cite{petersen2021deep}, which could better align with our risk-seeking evaluation pipeline. Additionally, incorporating the estimated remaining runtime of each VM and future VM demands \cite{mao2019learning} could further enhance performance. Predicting resource usage patterns may also help prevent performance inference by VMs hosted on the same PM. Furthermore, our current action design requires the agent to migrate VMs one at a time. Permitting the agent to swap multiple VMs simultaneously could simplify the identification of a feasible migration path. Overall, we hope our released datasets and RL environment will facilitate future research in this direction.

\section*{Acknowledgement}
We would like to thank our shepherd Mangpo Phothilimthana and anonymous EuroSys reviewers for their valuable comments and insightful feedback.

{\footnotesize \bibliographystyle{acm}
\bibliography{sample}}

\end{document}